\documentclass[sigconf]{acmart}

\usepackage{svg}
\usepackage{tikz}
\usepackage[flushleft]{threeparttable}
\usepackage{algorithm} 
\usepackage{algpseudocode} 

\usepackage{enumitem}
\usepackage{multirow}

\AtBeginDocument{%
  \providecommand\BibTeX{{%
    \normalfont B\kern-0.5em{\scshape i\kern-0.25em b}\kern-0.8em\TeX}}}

\acmConference[ICSE 2024]{46th International Conference on Software Engineering}{April 2024}{Lisbon, Portugal}

\begin{document}

\title{A New Frontier of AI: On-Device AI Training and Personalization}


\author{Jijoong Moon}
\email{jijoong.moon@samsung.com}
\orcid{0000-0003-0888-2143}
\authornote{The corresponding author.}
\affiliation{\country{}}

%% Device MLOps && NNStreamer
\author{Hyun Suk Lee}
\email{hs89.lee@samsung.com}
\affiliation{\country{}}

\author{Jiho Chu}
\email{jiho.chu@samsung.com}
\affiliation{\country{}}

\author{Donghak Park}
\email{donghak.park@samsung.com}
\affiliation{\country{}}

\author{Seungbaek Hong}
\email{sb92.hong@samsung.com}
\affiliation{\country{}}

%% Learning Framework
\author{Hyungjun Seo}
\email{hyungjun.seo@samsung.com}
\affiliation{\country{}}

\author{Donghyeon Jeong}
\email{dhyeon.jeong@samsung.com}
\affiliation{\country{}}

\author{Sungsik Kong}
\email{ss.kong@samsung.com}
\affiliation{\country{}}

\author{MyungJoo Ham}
\email{myungjoo.ham@samsung.com}
\affiliation{\country{}}

\author{}
\affiliation{%
  \institution{~\\\mbox{Samsung Research, Samsung Electronics}}
  \city{Seoul}
  \country{Republic of Korea}
}

\date{September 2023}

\renewcommand{\shortauthors}{Moon, et al.}

\begin{abstract}

Modern consumer electronic devices have started executing deep learning-based intelligence services on devices, not cloud servers, to keep personal data on devices and to reduce network and cloud costs.
We find such a trend as the opportunity to personalize intelligence services by updating neural networks with user data without exposing the data out of devices: \textit{on-device training}.
However, the limited resources of devices incurs significant difficulties.
We propose a light-weight on-device training framework, \textit{NNTrainer}, which provides highly memory-efficient neural network training techniques and proactive swapping based on fine-grained execution order analysis for neural networks.
Moreover, its optimizations do not sacrifice accuracy and are transparent to training algorithms; thus, prior algorithmic studies may be implemented on top of \textit{NNTrainer}.
The evaluations show that \textit{NNTrainer} can reduce memory consumption down to \textbf{1/20} (saving 95\%!) and effectively personalizes intelligence services on devices.
\textit{NNTrainer} is cross-platform and practical open-source software, which is being deployed to millions of mobile devices.

\end{abstract}


\maketitle

\section{Introduction}

We have witnessed the rapid proliferation of deep neural networks for a wide range of consumer electronics products in the industry.
Their intelligence services provide key features to consumer electronics: semantic segmentation~\cite{pal1993review} for smartphone cameras, super-resolution~\cite{yang2019deep} for TVs, object detection for robotic vacuums, image classification~\cite{krizhevsky2017imagenet,wu2015deep} for smart ovens, speech recognition~\cite{dahl2011context,hinton2012deep} for smartphones and TVs, and ASR~\cite{ling2015deep} and TTS~\cite{qian2014training} for real-time translations.
The quality of such services has become significantly important for consumer electronics; thus, a lot has been invested to develop and optimize deep neural networks.

Since the proliferation of intelligence services in consumer electronics, the need for adaptation to different environments and requirements of each individual user, ``personalization'', has arisen with additional technical difficulties~\cite{netflix_personalization}.
Because we use general data to train neural network models for the general public, not personal data for a personal model, for an individual user, the model may look over-parameterized or its quality of service is low compared to the size of the model.

To personalize intelligence services, we target to update models on devices with data available on devices: ``on-device training''.
Note that we do not need to train the whole models on devices; with personal data, we can update, fine-tune, or adjust models pre-trained by general data.
We may increase the accuracy of a user with the user's data.
We may add classes defined by a user to personalize the service for the user.
We may reduce latency and energy consumption for a user by reducing or skipping over-parameterized parts~\cite{7900006}.

Training models on devices instead of clouds provides significant advantages.
By running intelligence services on devices, ``on-device AI''~\cite{MITTechReview,ham2021nnstreamer}, we can save cloud operating costs (we have hundreds of millions of active mobile phones and consumer electronics deployed!), and we can keep personal data and privacy without exposing them externally.
By training on devices, not clouds, we can achieve the same advantages.
Besides, regulations and consumers' expectations on personal data and privacy are becoming more strict; e.g., the General Data Protection Regulation (GDPR)~\cite{voigt2017eu} makes it extremely difficult to gather personal activity records in clouds.
A frequently used device (e.g., a mobile phone) usually generates personal data continuously, which can enable continuous personalization with online or continual learning techniques~\cite{delange2021continual,thrun1995lifelong,shin2017continual} on the device.
Federated learning~\cite{li2020review,essalmi2010fully} usually requires on-device training mechanisms, too.

Limited resources of devices pose challenges to training on devices especially if the accuracy cannot be sacrificed for resources; i.e., business units usually prohibit such trade-off in the authors' affiliation.
For example, studies to reduce the memory overhead emerge as larger neural networks become popular: dynamic sparse reparameterization~\cite{mostafa2019parameter}, low precision training~\cite{narang2018mixed}, reduced batch sizes\cite{huang2019gpipe}, and gradient checkpointing~\cite{gradientchecking,chen2016training}, which address algorithmic aspects of models.
However, the system software aspect for training optimization is seldom studied, and we can improve significantly by addressing the structure and complexity of training software implementations.
We address the system software aspect (i.e., how memory and computation are allocated and scheduled) to optimize, which has been often overlooked in neural network frameworks.
Our approach does not sacrifice accuracy to conserve resources, which is often prohibited by applications.
Moreover, the algorithmic improvements mentioned above can be applied transparently on top of the proposed mechanism.

We address the execution orders of training procedures based on the observation that we can statically calculate computation and memory requirements from the model structure.
Thus, by controlling such execution orders, we can optimize resource utilization.
First, we divide a training session into fine-grained procedures and determine the life cycles of memory blocks assigned for procedures so that controlling execution orders of the procedures can determine the memory consumption accordingly.
Then, we identify possible execution orders for each neural network layer type and schedule them to minimize peak memory consumption.
Finally, we apply another optimization technique, Proactive Swap; i.e., we know when a buffer is read or written (clairvoyant!); thus, we can swap in and out proactively, minimizing its performance impacts.

Our contributions can be summarized as follows:
\begin{itemize}
\item We propose a highly memory-efficient on-device training technique without sacrificing accuracy and latency, exploiting the novel observation on the nature of neural network training, and Proactive Swapping.
The peak memory consumption is dramatically reduced, realizing training on embedded devices.
Moreover, the techniques are general; conventional machine learning frameworks may adopt them for servers, and neural network algorithmic optimization may be applied simultaneously.

\item We implement and release the proposed techniques as an open source, cross-platform, and commercialization-ready framework, \textit{NNTrainer}, which is already applied to actual products.

\item We evaluate \textit{NNTrainer} with various models and conventional frameworks, and show that it is highly efficient and effective.
We demonstrate that complex encoder-decoder models based on Tacotron2~\cite{shen2018natural} and Transformer~\cite{vaswani2017attention} can be personalized with sufficient batch sizes on mobile phones, which are deployed for products in early 2023.
\end{itemize}

\section{Related Work}\label{S:RelatedWork}

With larger neural networks being popular, studies on algorithms to consume less resources to train have been published, which usually trade-off between resources and accuracy with neural network algorithmic approaches.
A study~\cite{mostafa2019parameter} proposes dynamic sparse reparameterization to reduce the memory requirements by making the weight and activation values sparse during training.
It generates smaller model sizes that incur less peak memory consumption and computation with the sparsity by sacrificing the accuracy.
By adopting 16 bits float precision instead of 32 bits, ~\cite{narang2018mixed} reduces the memory and computation requirements, which also reduces the model size; however, it also sacrifices the accuracy.
The activation values occupy a significant part of memory and microbatching technique~\cite{huang2019gpipe} helps reduce the size of such values.
However, it alters the statistical properties of batch normalization, which affect the accuracy.
There are studies ~\cite{gradientchecking,chen2016training} to reduce memory consumption without sacrificing accuracy by storing the activation values partially and recomputing the activation values not stored; however, obviously, this increases the computation cost by about 30\%.
ZeRO-Offload~\cite{ren2021zero} and ZERO-Infinity~\cite{rajbhandari2021zero} propose software optimization mechanisms without sacrificing the accuracy by swapping the memory block to the secondary storage.
However, unlike \cite{ren2021zero,rajbhandari2021zero}, which swaps in and out reactively, the swapping mechanism of \textit{NNTrainer} swaps in and out according to the execution order; i.e., the swapping becomes proactive, which minimizes performance deterioration.
Another difference is that they focus on offloading the sliced chunks of activation or weight; thus, it is difficult for \cite{ren2021zero,rajbhandari2021zero} to support gradient clipping or non-trainable layers.

There are several popular software frameworks to train the neural network model: TensorFlow~\cite{199317} and PyTorch~\cite{paszke2017automatic}. 
TensorFlow is a popular open-source machine learning framework from Google.
It includes Keras and various tools for developers.
A lightweight variant of TensorFlow, called TensorFlow-Lite~\cite{tflite}, targets on-device AI, and training features have been enabled recently.
It is written in C/C++ rather than Python.
However, unlike \textit{NNTrainer}, developers cannot update model structures with the lightweight TensorFlow-Lite on device; the full TensorFlow is required for such tasks.

PyTorch~\cite{paszke2017automatic}, including Caffe2~\cite{jia2014caffe}, is another popular open-source machine learning framework from Facebook, which includes highly intuitive APIs.
It has an experimental lightweight variant, PyTorch Mobile, which runs mobile models converted from PyTorch models.
Note that PyTorch Mobile targets inference and does not support training.

TensorFlow, TensorFlow-Lite, and PyTorch consume an excessive amount of memory to train models (shown in \S\ref{S:Evaluation}); thus, consumer electronics cannot use them to train models.
For example, the quality assurance team of the authors' affiliation will drop in-house applications consuming an excessive amount of memory because the main memory may be crowded by a large number of applications preloaded~\cite{lim2023mobicom}.
Note that reducing the memory consumption is beneficial for conventional server-based machine learning; i.e., we may increase the batch size with less cost.

\section{Analysis}\label{S:Analysis}
\begin{figure*}[ht]
  \centering\includegraphics[width=0.90\textwidth]{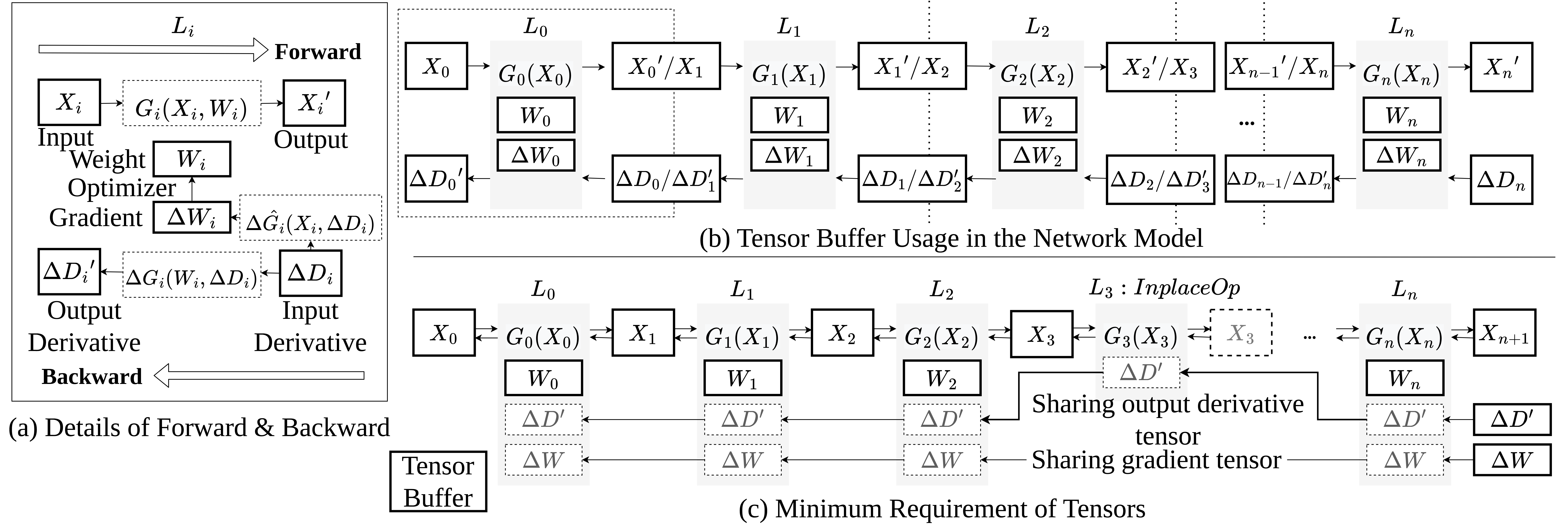}
  \caption{Memory buffer usage of forward and backward processes.}
  \label{FIG_LAYERS_FORWARDING_BACKWARDING}
\end{figure*}

%% \begin{table*}[tb]
%%   \centering
%%   \begin{threeparttable}

%%   \resizebox{\textwidth}{!}{
%%     \begin{tabular}{l c c c c c c c c c c c c c c c c c}
%%          & Total & C & C\_1 & C\_2 & C\_3 & C\_4 & C\_5 & C\_6 & C\_7 & C\_8 & C\_9 & C\_10 & C\_11 & C\_12 & D & D\_1 & D\_2\\
%%          &&&&&&&&&&&&&&&&&\vspace{-1em}\\
%%          \hline
%%     unknowns  & 14,939K & 1.7K & 36.9K & 73.8K & 147.5K & 295.1K & 590.0K & 590.0K & 1180.1K & 2359.8K & 2359.8K & 2359.8K & 2359.8K & 2359.8K  & 131.3K & 65.7K & 25.7K\\
%%          &&&&&&&&&&&&&&&&\vspace{-1em}\\
%%     (b) \tiny \scalebox{1.5}{(MiB)} & 591.0 & 68.6 & 134.5 & 50.9 & 68.2 & 27.5 & 38.2 & 38.2 & 22.0 & 35.6 & 35.6 & 23.0 & 23.0 & 23.0 & 1.8 & 1.0 & 0.5 \\
%%          &&&&&&&&&&&&&&&&\vspace{-1em}\\
%%     (c) \tiny \scalebox{1.5}{(MiB)}  & 163.7 & 1.5 & 33.7 & 8.6 & 17.3 & 5.3 & 10.7 & 10.7 & 6.8 & 13.6 & 13.6 & 10.4 & 10.4 & 10.4 & 0.7 & 0.3 & 9.7 \\
%%          &&&&&&&&&&&&&&&&\vspace{-1em}\\
%%          \hline
%%   \end{tabular}}
%%   \begin{tablenotes}
%%     \item \tiny \scalebox{1.5}{Not includes layers without unknows such as pooling, activation, etc. If optimizer is adam, about 100 MiB additional memory is required to save momentum.}
%%   \end{tablenotes}
%%   \end{threeparttable}
%%   \caption{Heap Memory Consumption of VGG16 (128 batch size, C:Conv2d, D:Dense)}
%%   \label{TBL_HEAP_MEMORY_CONSUMPTION}
%% \end{table*}
This section elaborates on the observation and analysis of the neural network training processes and their resource requirement and management.
Figure~\ref{FIG_LAYERS_FORWARDING_BACKWARDING} describes the memory usage for forward and backward processes.
Let's denote the network \(T=\{Y,G(X_0)\}\), where \(G(X_0)=G_n(G_{n-1}(...G_1(G_0(X_0))))\) as a sequence of layers and \(Y\) is label.

In a forward process of a layer, \(L_i\), in Figure~\ref{FIG_LAYERS_FORWARDING_BACKWARDING}.a takes \(X_i\) as an input, calculates \(G_i\), and saves the output, \({X_i}'\).
\(X_i\) and \({X_i}'\) are saved for the gradient calculation of backward operations.
Unlike a forward process, a backward process starts from the last layer, where the loss is calculated and propagated.
A backward process calculates the gradient \(\Delta W_i\) with the input derivative \(\Delta {D_i}\) and \(X_i\) saved in a forward process, and updates the weight \(W_i\) using an optimizer.
Then, it calculates the output derivative, \(\Delta {D_i}'\), to propagate the input derivative \(\Delta {D_i}\) to \(L_{i-1}\).

Depending on the layer type, a layer requires buffers for weight \(W_i\), gradient \(\Delta W_i\), input \(X_i\), and output \(G_i(X_i)\).
Usually, the sizes of such buffers depend on the input size and the depth of the neural network.
When the input size and the network configuration are determined, the amount of memory required for training can be roughly calculated.
Assume that there is a convolution 2D layer with the same padding, \(1 \times 1\) stride, and 64 filters with \(3\times3\) size, which takes an input image of \(32 \times 32 \times 3\).
Then, the input size is 0.39 MiB for 32 batch size (width $\times$ height $\times$ channels (3) $\times$ sizeof (float) $\times$ batch size),
and the output buffer size is 8.3 MiB (width $\times$ height $\times$ channels (64) $\times$ sizeof (float) $\times$ batch size).
We also need derivatives (\(\Delta D_i\), \(\Delta D_{i+1}\)), weight (\(W_i\)) and gradient (\(\Delta W_i\)) for backward process.
Therefore, we need about 16.6 MiB of heap memory to train a single layer.
For ``deep'' neural networks, we need multiple times, often over a hundred times, of such an amount of memory depending on the depth of the given neural network model.

Figure~\ref{FIG_LAYERS_FORWARDING_BACKWARDING}.b shows the memory buffers for neural network training.
It repeatedly reuses the memory buffer configuration of a single layer depicted in Figure~\ref{FIG_LAYERS_FORWARDING_BACKWARDING}.a, which is equivalent to dotted rectangles in Figure~\ref{FIG_LAYERS_FORWARDING_BACKWARDING}.b.
Conducting forward and backward processes for \(n\) layers requires memory space of about \(n/2\) times the memory space required for a single layer.
Because the output of \(L_i\), \({X_i}'\), and the input of \(L_{i+1}\), \(X_{i+1}\), represent the same data and layers do not modify input data, the two may share the same buffer instance.
Besides, we can reduce memory further as shown in Figure~\ref{FIG_LAYERS_FORWARDING_BACKWARDING}.c.
Although the output of each layer, \({X_i}'\), is stored for the backward process, its derivatives, \(\Delta D_i\) and \(\Delta D_i'\), are not required after the completion of the layer's backward process.
Therefore, memory space for the derivatives can be shared.
The same optimization can also be applied to the gradients because they are not required once the weight is updated with the gradients.
%Backward process of \(L_i\) uses input derivative \(\Delta D_i\) and input to the layer \(X_i\) to calculate output derivative \(\Delta {D_i}'\).
%Therefore, an input of a layer (\(L_i\)), \(X_i\), generally needs to be stored until the end of backward process including its previous layer, \(L_{i+1}\).

There are layers that may reuse input buffers, \(X_i\), as output buffers; e.g., activation layers.
For example, let \({X_i}'\) be the output of a sigmoid activation function, then, its derivative is \(\Delta {D_i}'={X_i}'(1-{X_i}')\).
During a backward process, computing \(\Delta {D_i}'\) requires \({X_i}'\), which is the output of a corresponding forward process, not the input of it.
Therefore, only one intermediate activation is required to store the output \({X_i}'\); i.e., an in-place computation shown as \(L_{3}: InplaceOp\) in Figure~\ref{FIG_LAYERS_FORWARDING_BACKWARDING}.c.
This allows for freeing the memory space storing inputs of activation layers.
Because activation layers are applied after most operations, including convolution and linear layers, this method reduces the memory requirement of inputs by almost half.
This can be applied to batch normalization as well.

%% Table~\ref{TBL_HEAP_MEMORY_CONSUMPTION} shows the ideal minimum memory requirements to forward and backward for vgg16 network with 128 batch size. The input and weight size are already known and we can identify how much minimum memory is required. About 591 MiB is required for the case Figure~\ref{FIG_LAYERS_FORWARDING_BACKWARDING} (b), but we can train with only 163 MiB for forwarding and backwarding as in Figure~\ref{FIG_LAYERS_FORWARDING_BACKWARDING} (c). Of course, there is additional memory used from the training framework, but it is very important to be able to calculate the minimum required memory like this when considering resource optimization.

%We can drop a significant part of buffers for inference to conserve memory space.
%For an arbitrary number of layers, allocating and alternating two buffers, \(X_i\) and \({X_i}'\), is sufficient.
%The two are temporal buffers for inferences, and we do not need to save \({X_i}'\) for backward process; thus, we do not need to save them.
%We may also remove the gradient buffer, \(\Delta W\), and the return derivative buffer, \(\Delta D'\).
%Besides, we do not need buffers for batch calculation to further reduce memory consu1mption. %% For the case of Table~\ref{TBL_HEAP_MEMORY_CONSUMPTION}, we only need the maximum input and output buffer size of memory, which is 0.5 MiB (\(32\times32\times64\times4\times2\)) and 59.7 MiB (\(14,941,604 \times 4\) for unknowns of all layers and that is about 60 MiB in total.

%% Introduction of Transfer learning, fine tuning, few-shot, recurrent?

\begin{figure}[t]
  \begin{center}
    \includegraphics[width=0.98\columnwidth]{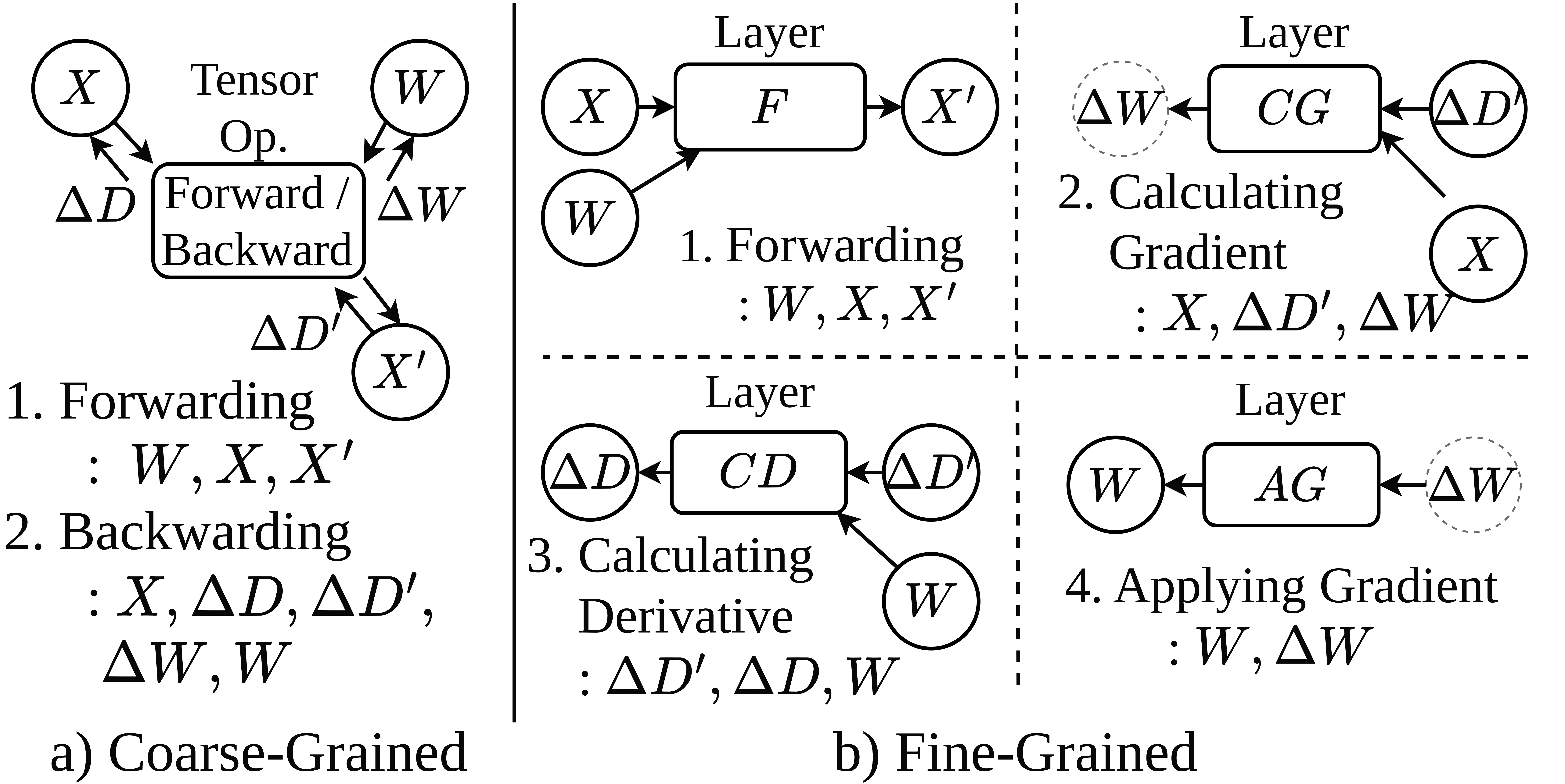}
  \end{center}
  \caption{Different granularity of training procedures. A gradient needs to be computed before the derivative; otherwise, \(X\) remains during derivative computation.}
  \label{FIG_TENSOR_LAYER}
\end{figure}

\begin{figure}[t]
  \begin{center}
    \includegraphics[width=0.98\columnwidth]{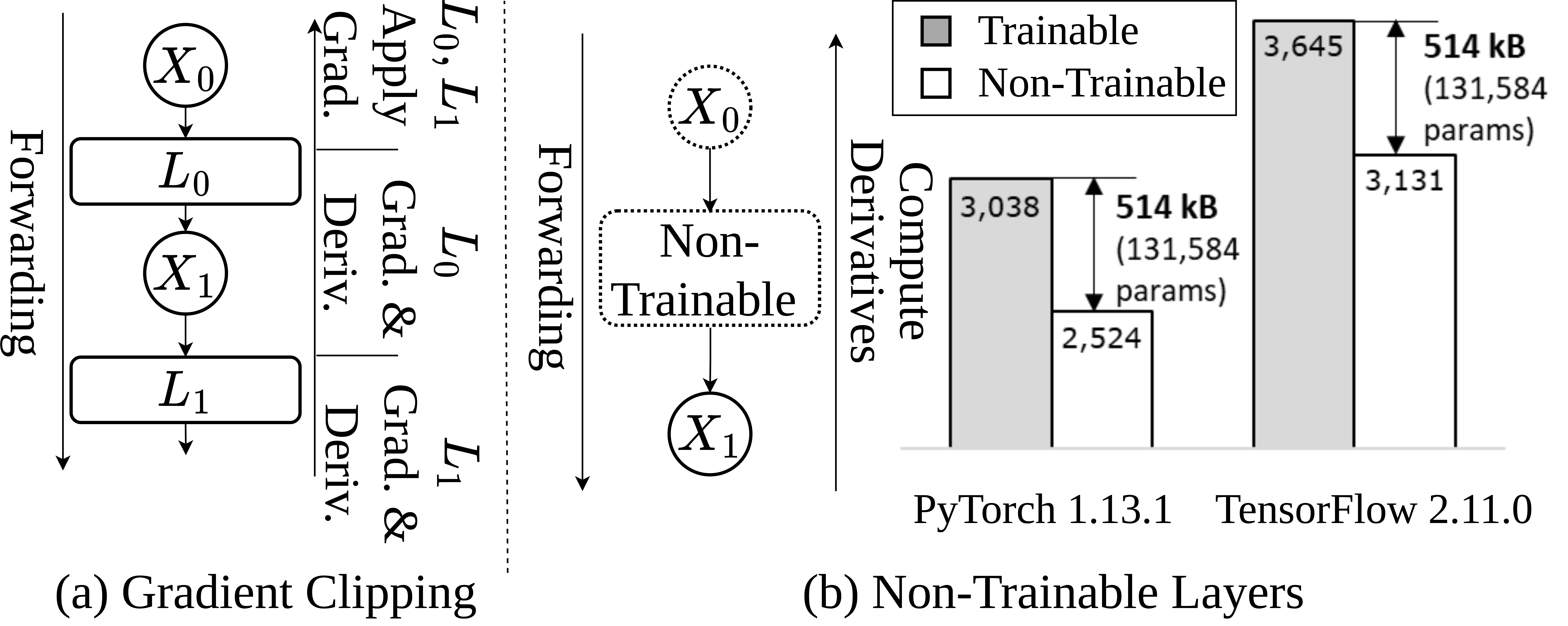}
  \end{center}
  \caption{Different types of training procedures.}
  \label{FIG_TRAINING_PROCEDURES}
\end{figure}

From the analysis, we find that different granularity of training procedure is possible, reducing the memory consumption with its own pros and cons as in Figure~\ref{FIG_TENSOR_LAYER}. 
Coarse-grained procedure as in conventional training frameworks (TensorFlow and PyTorch) costs less to develop a new layer because only its forward process is required to be implemented with Automatic Differentiation~\cite{baydin2018automatic}; however, it is difficult to systematically find the minimum memory requirement.
%We can reuse significant parts of tensors based on the order of the input tensors of a given operator.
%However, it is extremely difficult to find the order of tensor operations because they vary according to how each developer implements a forward process and it is also difficult to handle an automatic differentiation engine simultaneously.
On the other hand, fine-grained training procedures can clearly distinguish such as \textit{forwarding}, \textit{calculating gradient}, \textit{calculating derivative}, and \textit{applying gradient}; thus, the activation and derivative memory, \(X\) and \(\Delta D\) are not required at the same time, so that we can minimize the memory consumption further.
\textit{NNTrainer} has fine-grained training procedures implemented by assigning the execution order (\textit{EO}) to each procedure, and it has scheduling algorithms(\S\ref{S:Design}) based on \textit{EO}.
With the scheduling algorithms, \textit{NNTrainer} overwrites invalid memory buffers such as \(X\) after computing gradients to avoid de-allocating and re-allocating buffers.
It also helps to reduce I/O interactions during the memory-to-storage swapping. Prior art~\cite{gradientchecking, chen2016training, shah2020memory} mainly focuses on managing the activation memory computed at the forwarding procedure. Unlike prior art, we find more opportunities during the backward procedures and propose a highly efficient memory optimization scheme without sacrificing accuracy. Therefore, previous works can be applicable on top of our work.

The resource optimization schemes should be applicable for various scenarios including examples in Figure~\ref{FIG_TRAINING_PROCEDURES}.
The proposed mechanism supports a mixed network of trainable layers and non-trainable layers, which is common in on-device AI personalization~\cite{torrey2010transfer, lu2015transfer, tan2018survey}.
In non-training layers, forwarding processes do not need to store \(X_0\) because of skipping gradient-related processes shown in Figure~\ref{FIG_TRAINING_PROCEDURES}.b.
Although conventional frameworks support non-trainable layers, they still allocate memory for activation inputs, \(X_0\).
We evaluate with a fully connected layer having 131,584 parameters (514 kiB of float32) in the middle of a model with GPUs and confirm that the memory of gradient is reduced as in Figure~\ref{FIG_TRAINING_PROCEDURES}.b.
Moreover, the re-ordering of fine-grained training procedures enables the Gradient Clipping~\cite{zhang2020improved} easily as in Figure~\ref{FIG_TRAINING_PROCEDURES}.a.
Therefore, the fine-grained training procedure is more robust and efficient for supporting various training processes and resource utilization optimizations.

\section{Design}\label{S:Design}

\begin{figure}[t]
  \begin{center}
    \includegraphics[width=0.97\columnwidth]{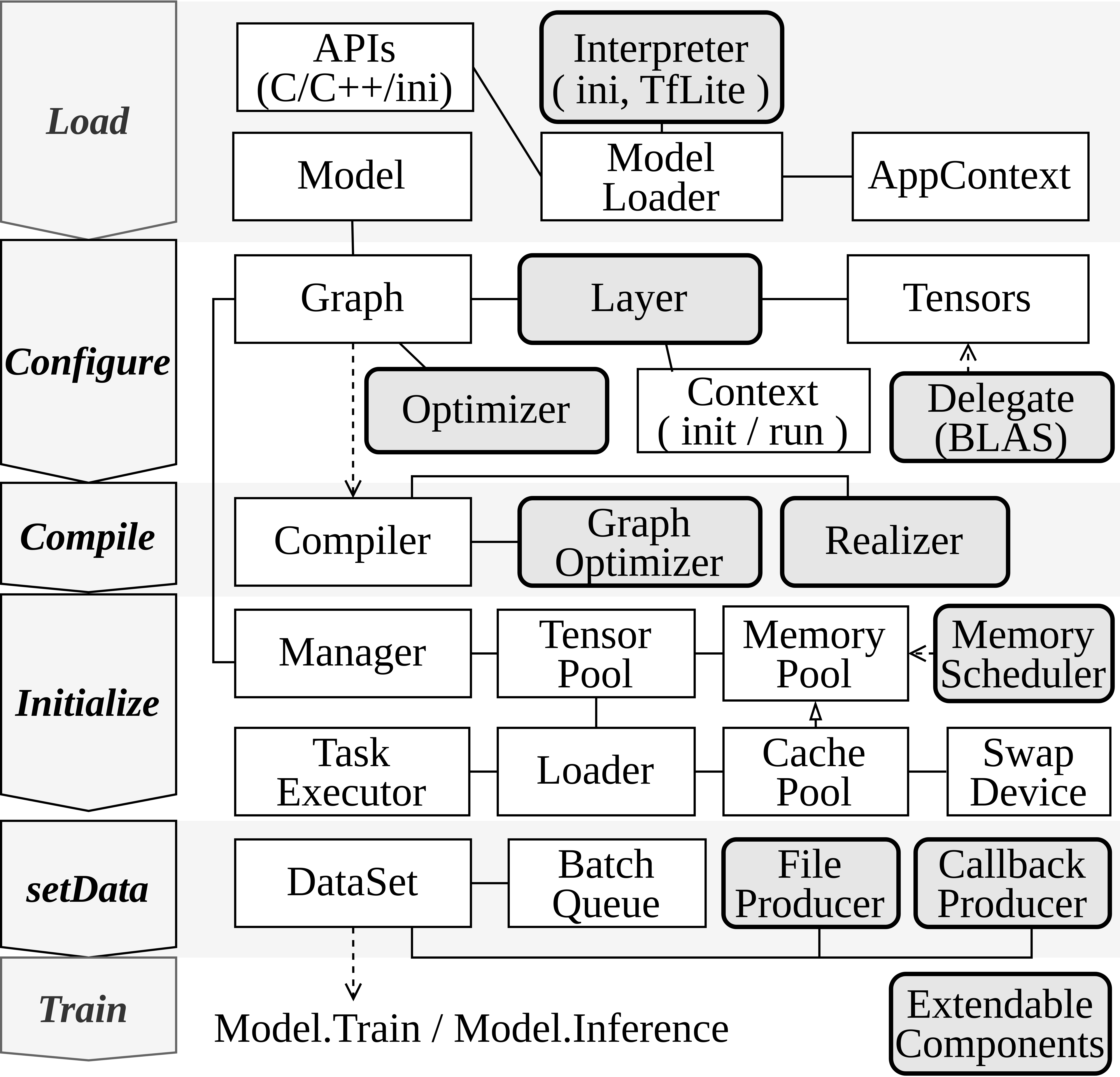}
  \end{center}
  \caption{Abstract architecture of \textit{NNTrainer}.}
  \label{FIG_SYSTEM_OVERVIEW}
\end{figure}

This section describes the design and implementation of \textit{NNTrainer}.
\textit{NNTrainer} is a modular framework written in C++ with many user-extensible components, whose structure is shown in Figure~\ref{FIG_SYSTEM_OVERVIEW}.
It is cross-platform; we release to Ubuntu, Tizen, Android, and Windows.

The training processes of \textit{NNTrainer} can be categorized as \textit{Load}, \textit{Configure}, \textit{Compile}, \textit{Initialize}, \textit{setData}, and \textit{Train}, in the order we elaborate.

\textbf{Load}: the output of \textit{Load} is an instance of Model, which has all the information of a neural network model, including hyper-parameters, and orchestrates all the components and processes.
Users may describe a neural network as an instance of Model with an initialization file~\cite{iniparser} or APIs in C, C++, .NET, and Web.
The interpreter in Model Loader parses model descriptions and generates intermediate graph representations.
Developers may extend Interpreters to support model files of other frameworks; we provide an extension for TensorFlow-Lite~\cite{199317}.
The output of \textit{Load} describes layers of the model as tuples of \texttt{\small{[<Layer type>, <Properties (key, value)>]}}.

\textbf{Configure} creates layer objects with those tuples, and the layer objects construct a Graph.
We construct Graph and Layer instances with delayed allocation based on the information provided by Context, which has parameters of Graph and Layer instances.
It conserves memory by delaying Tensor buffer allocations until the buffers are actually required.

\textbf{Compile}: before \textit{Initialize}, Layer Context exists as an initContext instance storing the information as strings.
Later, initContext becomes runContext by \textit{Initialize} process, converting string descriptions to Tensor objects.

It is lighter and more efficient to conduct the \textit{Compile} process with initContext; it handles strings, not Tensor instances.
Each Layer subclass provides forward and backward functions that calculate gradients and derivatives, respectively.
Each Optimizer subclass provides a function applying gradients.
Users can implement new layers and optimizers by adding these subclasses.

For higher memory efficiency, we need to manage Tensor data independently; thus, we have separated specifications (dimensions and status) and data stored in buffers.

\begin{table}[tb]
  \centering
  \begin{threeparttable}
    \resizebox{\columnwidth}{!}{
    \begin{tabular}{l l}
      Realizer & Role\\
      &\vspace{-1em}\\
      \hline
      Flatten & Identify and create a flatten layer\\
      Activation & Identify and create an activation layer\\
      BatchNorm & Identify and create a batch normalization layer\\
      Multi-Out & Identify and make connection to layers\\
      Loss & If loss is cross entropy, remove the activation\\
      Concat & Identify inputs and create Concatenate layer\\
      Recurrent & Unroll the graph if there is a loop\\
      Slice & Create sub-graph network in the backbone model\\
      Input & Identify and create an input layer\\
      &\vspace{-1em}\\
      \hline      
    \end{tabular}}
  \end{threeparttable}
  \caption{Default Realizer subclasses.}
  \label{TBL_REALIZER}
\end{table}

The Model's Graph, created by \textit{Configure} process, can operate after Realizer of Compiler applies lowering operations; i.e., the Realizer adds layers or changes the order based on the analysis of graph structures.
%Such operations make neural network model more concrete.
Table~\ref{TBL_REALIZER} shows the default Realizer subclasses of \textit{NNTrainer}.
\textit{Compile} process computes connection and ordering with the provided Graph Optimizer subclasses, which are critical for memory saving.
Users may extend both Graph Optimizer and Realizer by adding subclasses.
%%As explained later (\S\ref{execution_order}), memory consumption significantly varies per different ordering; thus, an ordering algorithm of \textit{Graph Optimizer} and \textit{Realizer} is extremely important for resource utilization.

%%Actual Tensor objects are not created, yet, in the previous processes, but only the information required to create Tensors is stored in \textit{initContext}.
\textbf{Initialize}: in \textit{Initialize} process, a finalize method of a Layer subclass is executed while visiting each Layer instance in Graph, and Tensor instances are requested to Tensor Pool.
Because specification and data of Tensor are managed independently, \textit{NNTrainer} manages memory separated by Tensor Pool and Memory Pool.
%%Although a Tensor instance is requested, its memory space is not yet allocated.
After Tensor Pool creates all requested Tensors instances throughout the Model, Memory Planner plans with Tensor instances given by Tensor Pool, computes the offset of the Memory Pool for each Tensor instances, and assigns the memory for Tensor data.
\textit{NNTrainer} provides the Memory Swap using Cache Pool and Loader.
Loader creates an independent Task Executor and monitors the existence of the tensor data in memory.
It loads and unloads proactively using execution order (\textit{EO}) to hide the data transfer overhead (\S\ref{S:Analysis}). 
%%Memory Planner is also extendable to support various management algorithms.

\textbf{setData}: to generate batch-sized data for training in \textit{setData} process, \textit{NNTrainer} provides DataSet, and users can provide training data using DataProducer, which is extendable.
DataProducer generates data for training and accumulates the data in the Batch Queue up to the batch size.
%File and Callback Producer are provided.
After the described processes, we are finally ready for \textbf{Train} process.

%% In view of the neural network structure, it can be largely divided into CNN-based feed-forward network and recurrent network, and the they are very differnt in terms of implementation. The recurrent network includes time sequence iterations of sub-network or layer, so that, the spliting and manipulating the network structure are required. To do this, \textit{NNTrainer} provides various neural network graph control layers such as, concat/addition, multi-input/output, split layers, and time unrolling.

In terms of resources, the uniqueness of \textit{NNTrainer} over the conventional is that Tensors in \textit{NNTrainer} are prioritized based on \textit{EO} and \textit{in-place operation}.
This allows Memory Planner to maximize the reusability of Memory Pool; thus, reducing memory consumption.

\subsection{Temporal and Spatial Relation} \label{execution_order}

\begin{table}[tb]
  \centering
  \begin{threeparttable}
    \resizebox{\columnwidth}{!}{
    \begin{tabular}{l l}
      Temporal Specification & fine-grained processes\\
      &\vspace{-1em}\\
      \hline
      Forward (\textit{F}) & forward\\      
      Compute Gradient (\textit{CG}) & compute gradient\\
      Compute Derivative (\textit{CD}) & compute derivative\\            
      Apply Gradient (\textit{AG}) & apply gradient \\
      Backward (\textit{B}) & compute gradient, derivative and apply gradient\\
      Iteration (\textit{I}) & reset after iteration\\
      Max (\textit{M}) & always available \\
      &\vspace{-1em}\\
      \hline      
    \end{tabular}}
  \end{threeparttable}
  \caption{Temporal relations.}
  \label{TBL_TENSOR_LIFE_SPAN}
\end{table}

During training, intermediate activation accounts for more than 90\% of the total memory consumption~\cite{cai2020tinytl}, and forward processes decide intermediate activation values, which are saved for backward processes to use.
Thus, if \textit{EOs} effectively exploit orders between processes, we can use memory more efficiently as in Figure~\ref{FIG_TENSOR_LAYER}.b.
To achieve this, \textit{NNTrainer} divides the entire training process into \textit{Forward}, \textit{Compute Gradient}, \textit{Compute Derivatives}, and \textit{Apply Gradient}, and configures \textit{EOs} of each layer.
It is straightforward to determine the orders of layers; however, memory scheduling and swapping require more information.
%For example, a Tensors instance can be overwritten without unloading if it is not valid anymore, and in-place operations such as activation and reshape may waste memory consumption significantly if not handled accordingly.

To address memory scheduling, \textit{NNTrainer} defines \textit{temporal relations} (Table~\ref{TBL_TENSOR_LIFE_SPAN}) and \textit{spatial relations} (Table~\ref{TBL_TENSOR_SHARING}) in Tensor specification.
Spatial relation is assigned automatically during Initialize process by analyzing the network graph.
If a buffer for intermediate activation is requested, \textit{Create(C)} is assigned by default.
However, if its previous layer is an in-place operation, \textit{Modify View (MV)} or \textit{Read-Only View (RV)} is assigned, determined by the behavior of the given layer for the memory buffer, and if the previous layer is ReLU or sigmoid, \textit{NNTrainer} assigns \textit{Modify View (MV)}.
If the examined layer is flatten or reshape, which modifies the dimensions of tensors, not their contents, then \textit{Read-Only View (RV)} is assigned.
\textit{Extend (E)} creates tensor sharing everything; e.g., weight tensors of an unrolled sub-graph for the recurrent network.

%%\textit{NNTrainer} provides several options to define tensor lifespan as in Table~\ref{TBL_TENSOR_LIFE_SPAN}. For example, a Tensor storing weight has max lifespan, which is valid always and cannot be deallocated during the entire training process.

\begin{table}[tb]
  \centering
  \begin{threeparttable}
    \resizebox{\columnwidth}{!}{
    \begin{tabular}{l l l}
      Dependency & Spatial Relation & Description\\
      &\vspace{-1em}\\
      \hline
      \multirow{2}{*}{Independent} & Place-Holder (\textit{P}) & Holding the external memory\\      
       & Create (\textit{C}) & Allocate new Tensor\\
       \hline
      \multirow{2}{*}{Memory sharing} & Modify View (\textit{MV}) & Data changes\\
       & Read-Only View (\textit{RV}) & No data change \\
       \hline
      Tensor sharing & Extend (\textit{E}) & Share everything\\
      &\vspace{-1em}\\
      \hline      
    \end{tabular}}
  \end{threeparttable}
  \caption{Spatial (inter-layer) relations.}
  \label{TBL_TENSOR_SHARING}
\end{table}

\begin{algorithm}[t]
    \scriptsize{}
	\caption{Compute Execution Order.} \label{alg:exec_order}
	\textbf{Input:} Layer List: ${L = \{L_0, L_1,..., L_{N-1}\}}$
	\begin{algorithmic}[1]
	    \State $EO_{max}=N\times4$
	    \State Initialize $T_{map}$
	    \For{$i=0,1,\ldots,N-1$}
	        \State $EO_{F}=i$
	        \State $EO_{CG}=N+i\times3$
                \State $EO_{AG}=EO_{CD}+1$         
            \If{Gradient Clipping}
                \State $EO_{CG}=N+i\times2$
                \State $EO_{AG}=N\times3+i$
            \EndIf    
            \State $EO_{CD}=EO_{CG}+1$
            \If{Non-Trainable}
                \State $EO_{CG,CD,AG}=N+i\times3$
                \Comment{Set $EO_{L_i}$ same \& run compute derivatives only}
            \EndIf
            \State $EO_{L_i}=\{EO_F,\ EO_{CG},\ EO_{CD},\ EO_{AG}\}_{Layer\ Type}$           
            \For{$t_r \in T_{L_i}$} \Comment{$T_{L_i}$ is Tensor list requested in $L_i$}
                \State $t_r = T_{map}.get(t_r)$ \textbf{or} $T_{map}.insert(t_r)$
                \State Add $EO \in EO_{L_i}$ to $t_r$ w.r.t Temporal  of $t_r$
            \EndFor
	    \EndFor
	    \State Sort $T_{map}$ by ascending with $min(EOs\ of\ T_i \in T_{map})$
	    \For{$T_i \in T_{map}$}
	        \State $CM_{T_i}$ = Tensor create mode of ${T_i}$
	        \If{$CM_{T_i} == MV_{T_j}$} \Comment{$T_{j}$ is target Tensor}
	            \If{$min(EOs\ of\ T_i) \geq max(EOs\ of\ T_j)$}
	               \State Merge $T_i$ to $T_j$ including $EOs$
	            \EndIf
	        \ElsIf{$CM_{T_i} == RV_{T_j}$ \textbf{or} $E_{T_j}$}
	            \State Merge $T_i$ to $T_j$ including $EOs$
	        \EndIf
	    \EndFor
	\end{algorithmic} 
\end{algorithm}
%\begin{figure}[t]
%  \begin{center}
%    \includegraphics[width=0.98\columnwidth]{figures/exe_order_fc.png}
%  \end{center}
%  \caption{Execution order and temporal-spatial relation (Case A). Refer to %Figure~\ref{FIG_LAYERS_FORWARDING_BACKWARDING}, Table~\ref{TBL_TENSOR_LIFE_SPAN}, and %Table~\ref{TBL_TENSOR_SHARING} for the notations.}
%  \label{FIG_EXE_ORDER_FC}
%\end{figure}

\begin{figure}[t]
  \begin{center}
    \includegraphics[width=0.98\columnwidth]{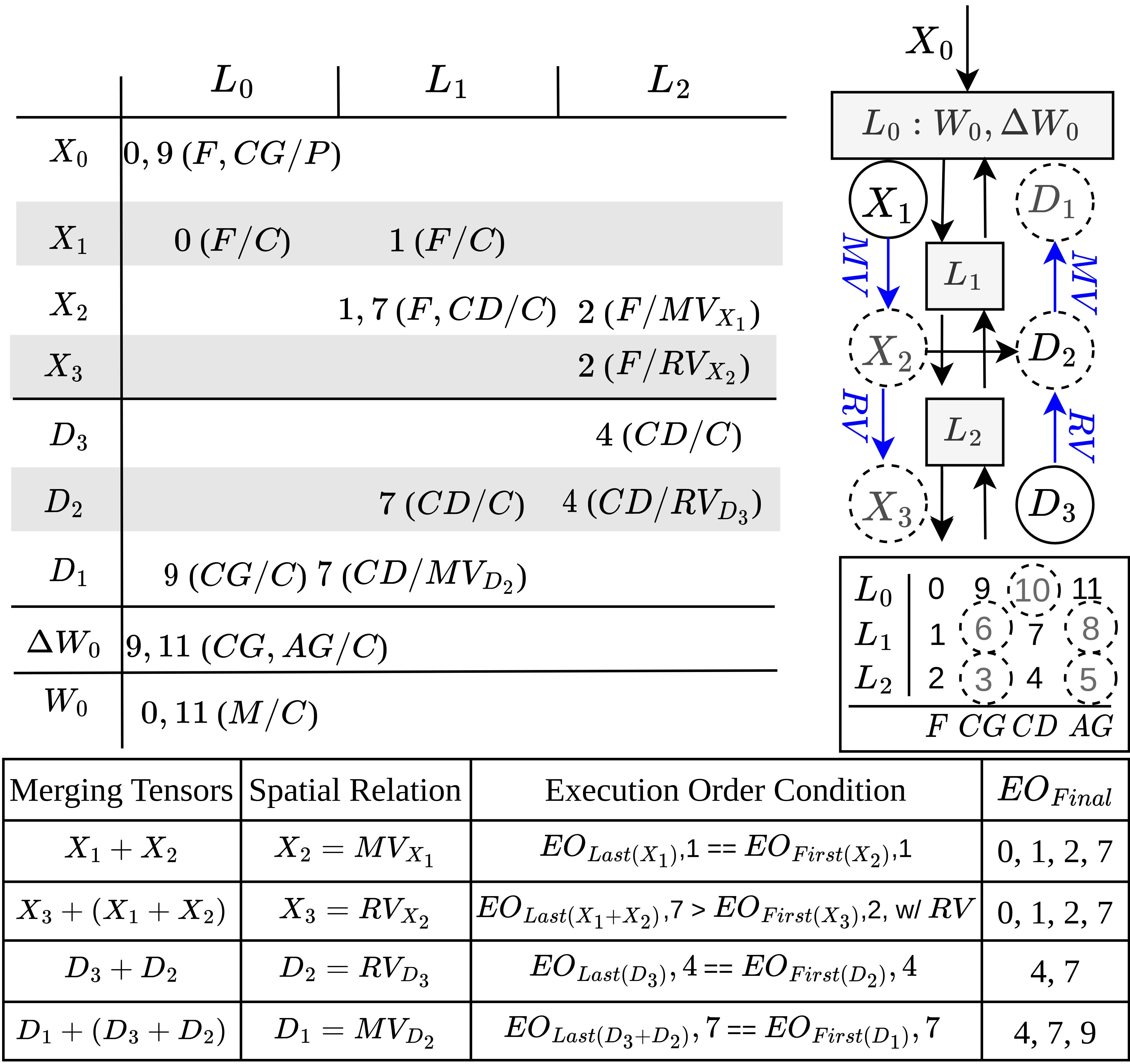}
  \end{center}
  \caption{Execution orders and temporal-spatial relations of an example model, where only \(X_0\), \(X_1\), \(D_3\), \(\Delta W_0\), and \(W_0\) are required. Refer to Figure~\ref{FIG_LAYERS_FORWARDING_BACKWARDING}, Table~\ref{TBL_TENSOR_LIFE_SPAN} and \ref{TBL_TENSOR_SHARING} for the notations.}
  \label{FIG_EXE_ORDER_FL}
\end{figure}

%Developers may use Tensor Lifespan and Tensor Create Mode in different ways depending on the network type and structure.
Algorithm~\ref{alg:exec_order} describes how \textit{EO} is defined using temporal and spatial relations of tensors.
Figure~\ref{FIG_EXE_ORDER_FL} shows how Algorithm~\ref{alg:exec_order} works for an exemplar model of three layers, where \(L_1\) is a sigmoid activation and \(L_2\) is a flatten layer.
In the figure, \(0,9(F,CG/P)\) denotes that the temporal relation is \(F\) and \(CG\), the spatial relation is \(P\), and the \textit{EOs} are 0 and 9.
It sequentially assigns \textit{EOs} starting from \(L_0\), and the result is shown in the bottom-right square in the figure. 
 %First, 0 and 7 are assigned to \(L_0\)'s \(F\), \(CG\) and \(CD\) is not included for Input Layer. 1, 5, and 6 are assigned to \(L_1\)'s \textit{EOs} and 2, 3, and 4 are for \(L_2\).

%Because \textit{CG} and \textit{CD} operate in the backward of \(L_2\), 3 and 4 are assigned respectively.
%Then, 5 and 6 are given to \textit{CG} and \textit{CD} in the same way for \(L_1\).
%Finally, 7 is given to \textit{CG} of \(L_0\) and \(L_0\) does not have \textit{CD}.
In \(L_0\), requested tensors are \(X_0\), \(X_1\), \(D_1\), \(\Delta W_0\), and \(W_0\).
Because the temporal relations of \(X_0\) are \(F\) and \(CG\), their \textit{EOs}, 0 and 9, are assigned to \(X_0\).
Then, set 0 (\(F\) of \(L_0\)) for \(X_1\) and set 9 (\(CG\) of \(L_0\)) for \(D_1\).
This procedure is repeated for each Layer, \(L_i\): line 3 to 20 in Algorithm~\ref{alg:exec_order}.
%After \textit{EOs} are determined for all layers, \textit{EOs} of each Tensor are combined together as shown in Figure~\ref{FIG_EXE_ORDER_FC}.
After merging output tensors, \(X_{i-1}'\) and \(X_i\) in Figure~\ref{FIG_LAYERS_FORWARDING_BACKWARDING}, we can calculate theoretical memory requirement of the given model, which provides the basis for peak memory consumption comparison in Evaluation.

%% but at this time, add it to the existing Tensor's execution order. Since the life span of \(X_1\) is forwarding and compute gradient, we add 1 (forwarding of \(L_1\)) and 5 (compute gradient of \(L_1\)). At his time, if there is an order larger than the max execution order given in \(L_1\) among the existing \(X_1\) execution orders, it cannot be added and new Tensor object is required. As you can see, the greater execution order means that it will be used later layer and it cannot be reused in \(L_1\). Here, \(X_1\) of \(L_0\) and \(X_1\) of \(L_1\) may be reused. Therefore, the execution order of \(X_1\) becomes 0, 1, 5.

%%\begin{figure}[t]
%%  \begin{center}
%%    \includegraphics[width=0.98\columnwidth]{figures/exe_order_ac.png}
%%  \end{center}
%%  \caption{Execution order of model B (activation \(L_1\)), where \(D_1\) and \(X_2\) are not allocated.}
%%  \label{FIG_EXE_ORDER_AC}
%%\end{figure}

%In general, finding the best \textit{EO} sequence is not trivial with complex network structures.
Spatial relations appear between an in-place layer and its adjacent layer as in Figure~\ref{FIG_EXE_ORDER_FL}.
%Due to the characteristics of activation layers, we do not need to calculate gradients to update a weight.
%Thus, \(W_1\) and \(\Delta W_1\) for the Layer \(L_1\) are not requested.
Then, the input, \(X_1\), and output, \(X_2\), of \(L_1\) can be Memory Sharing (Table~\ref{TBL_TENSOR_SHARING}) with changing data, and \(X_2\) can be marked as Modify View of \(X_1\) (\(MV_{X_1}\)).
Read-Only View (\(RV\)) of \(X_2\) (\(RV_{X_2}\)) can be given to \(X_3\).

For a given spatial relation of \(L_1\), let's call \(X_1\) a target Tensor and \(X_2\) a merged Tensor.
Similarly, \(X_2\) is a target Tensor for a merged Tensor \(X_3\) for a spatial relation of \(L_2\).
%After \textit{EO}s for Tensors are combined, Tensor Create Mode needs to be determined; e.g., \textit{MV} of \(X_2\).
%First, we start from \(L_0\) as in forward process.
%The current \textit{EO} of \(X_2\), which is Modify View of \(X_1\), is 1, 2, 3, and 6, and \(X_1\) is 0 and 1. 
When applying spatial relation, only the largest order of the target Tensor and the smallest order of the merged Tensor are compared: line 25 of Algorithm~\ref{alg:exec_order}.
The largest order of \(X_1\), 1, is equal to the smallest order of the merged Tensor, \(X_2\), 1; thus, these Tensors and orders can be merged as 0, 1, 2, 7.
If the largest order of the target Tensor is greater than the smallest order of the merged Tensor, the integrity of the target Tensor cannot be guaranteed because the target Tensor is accessed after the merge; thus, it cannot be combined and a new Tensor needs to be created.
Next, the largest order of \(X_1 + X_2\) is greater than the smallest order of the merged Tensor, \(X_3\), 2; however, it can still be merged because the integrity of data is guaranteed with \textit{RV}.
This reduces memory requirement by removing another intermediate activation as described in Figure~\ref{FIG_LAYERS_FORWARDING_BACKWARDING}.

\subsection{Memory Pool and Memory Planner}

\begin{figure}[t]
  \begin{center}
    \includegraphics[width=0.98\columnwidth]{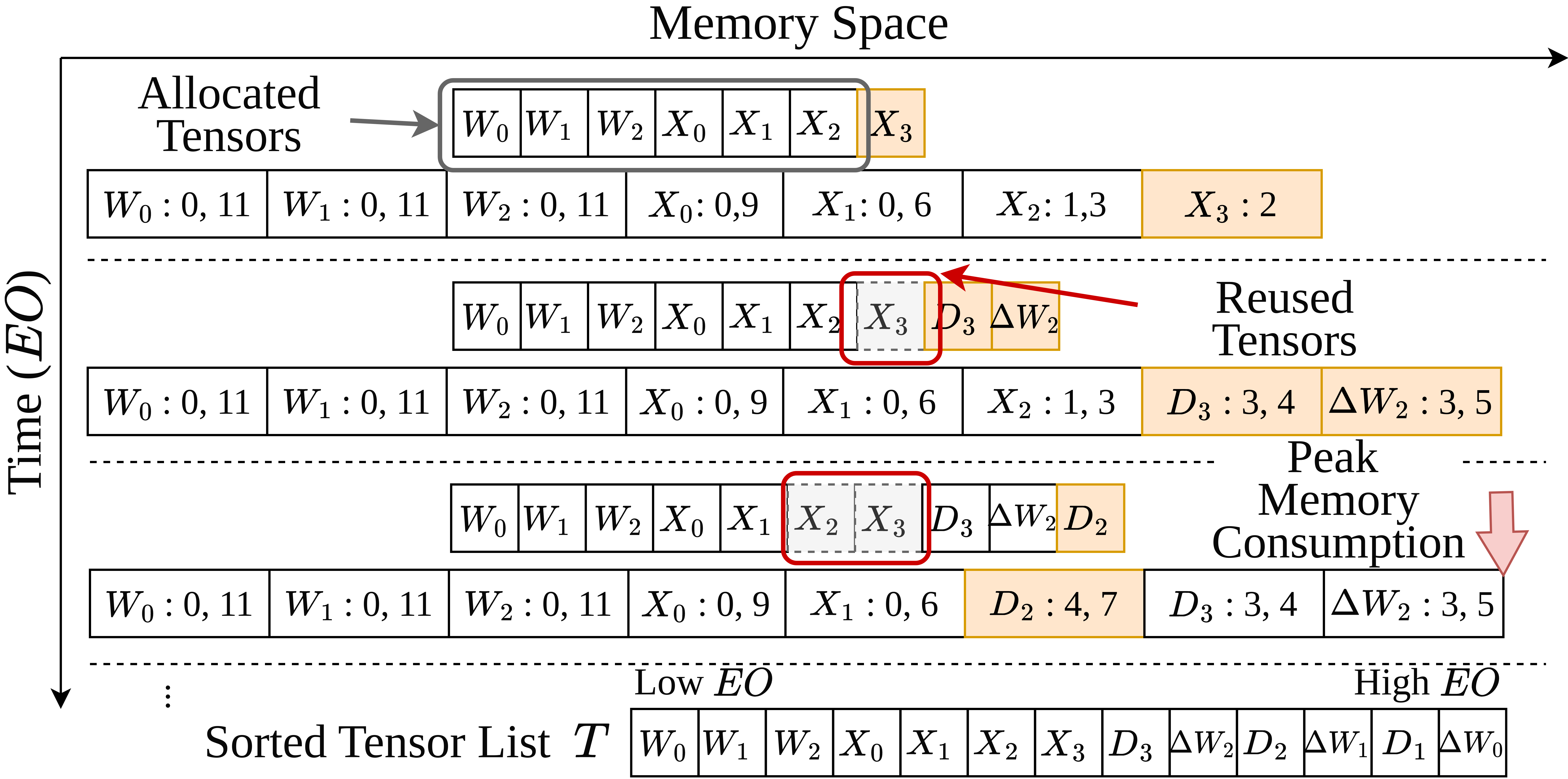}
  \end{center}
  \caption{Memory planning without spatial relation for three linear layers.}
  \label{FIG_MEM_PLANNER_FC}
\end{figure}

With \textit{EOs} assigned, Memory Planner allocates buffers from Memory Pool with the planner algorithm, Algorithm~\ref{alg:planner}.
Algorithm~\ref{alg:planner} is simple sorting-based; an algorithm minimizing fragmentation for higher utilization is future work.
Figure~\ref{FIG_MEM_PLANNER_FC} shows how Algorithm~\ref{alg:planner} works for a case without spatial relations as a simple example, where the four iterations of the \textbf{for}-loop, \(i = [7 ... 10]\), are shown, reallocating a few tensors to share and reuse the memory space.

\begin{algorithm}[t]
    \scriptsize{}
	\caption{Simple sorting-based Memory Planner.} \label{alg:planner}
	\textbf{Input:} Tensor List: ${T = \{T_1, T_2,..., T_N\}}$
	\begin{algorithmic}[1]
	    \State Sort T by ascending order based on $min(EOs\ of\ each\ T_i)$
	    \If{ $min(EOs\ of \ T_i)$ == $min(EOs\ of\ T_j)$}
	        \State Sort by descending order based on $max(EOs\ of\ T)$
	    \EndIf
        \For {$i=1,2,\ldots,N$}  \Comment{${T}$ is Sorted, now}
		    \State ${EO_{i}}^{min}$ = $min\left(EO\ of\ {T_i}\right)$
		    \State $last = i$
            \For {$j=i-1,\ldots, 1$}
                \State ${EO_{j}}^{max}$ = $max\left(EO\ of\ {T_j}\right)$
                \If{${EO_{j}}^{max} < {EO_{i}}^{min}$}
                     \State $last = j$ 
                \EndIf
            \EndFor

            \If{${last} == {i}$}
                 \State Calculate offset for ${T_{i}}$ and Assign to ${T_{i}}$ data              
            \Else
                \State Assign data(offset) of ${T_{last}}$ to ${T_{i}}$ data(offset)
            \EndIf
    	\EndFor
	\end{algorithmic} 
\end{algorithm}

%First, the planner lists all Tensors in the order of \textit{EO} values.
%It sorts in ascending order based on the smallest \textit{EO} of the Tensor.
%If the smallest \textit{EO}s are same, then, it sorts in descending order based on the largest \textit{EO} of Tensors.
%After Tensors are sorted, memory blocks are sequentially allocated for Tensor data.
%At this point, the largest order of the previously allocated Tensors are compared one by one in opposite direction with the smallest order of the Tensor to be allocated.
%If the largest order is smaller than the smallest order, we can remove the Tensor, so that we can reuse. We keep remove the Tensors until there is no smaller cases, the address of last available Tensor data is assigned to the Tensor data to be allocated.
%After reviewing until there are no smaller cases, the address of the last available memory block  is assigned to the Tensor data.

For each Tensor data, we calculate the memory offset to assign.
%The offset is same as the size of Tensor.
In Figure~\ref{FIG_MEM_PLANNER_FC}, when we calculate the offset of \(W_1\) after calculating the offset of \(W_0\), because there are no assigned Tensors whose largest \textit{EO} is less than the smallest \textit{EO} of \(W_1\), a new offset is calculated and assigned to \(W_1\).
It keeps assigning new offsets until \(X_3\) as shown in the first row of Figure~\ref{FIG_MEM_PLANNER_FC}.
As in the second row of the figure, when we calculate for \(D_3\), the largest \textit{EO} of \(X_3\), 2, is less than the smallest \textit{EO} of \(D_3\), 3.
It means that \(X_3\) can be reused; thus, the same offset is assigned to the data of \(D_3\) as in line 17 of Algorithm~\ref{alg:planner}.
To assign \(\Delta W_2\) in the next step, a new offset is required because there is no Tensor whose largest \textit{EO} is smaller than the smallest \textit{EO} of \(\Delta W_2\).
%Next, when \(D_2\) is allocated, since the largest order of \(X_2\), 3, is smaller than the smallest order of \(D_2\), 4, the memory block address is assigned to \(D_2\). 
Memory planning is completed when calculating offsets and assigning Tensors are completed.

%%\begin{figure}[t]
%%  \begin{center}
%%    \includegraphics[width=0.98\columnwidth]{figures/memory_pool_ac.png}
%%  \end{center}
%%  \caption{Memory Planning with Case B.}
%%  \label{FIG_MEM_PLANNER_AC}
%%\end{figure}

%Figure~\ref{FIG_MEM_PLANNER_AC} shows the memory planning example with Model B.
%Unlike Model A, there is a Tensor that cannot be reused.
%For example, \(D_2\) causes fragmentation when \(\Delta W_0\) is assigned as shown in the last row of Figure~\ref{FIG_MEM_PLANNER_AC}.
%Therefore, a planning algorithm that can minimize or resolve fragmentation is required; however, the implementation of such an algorithm is future work for the next releases.
%This is also works well for the Model C in Figure~\ref{FIG_EXE_ORDER_FL}.

A major advantage of this method is that we can calculate the peak memory consumption beforehand as shown in Figure~\ref{FIG_MEM_PLANNER_FC}, which is equal to the ideal memory requirement in Figure~\ref{FIG_LAYERS_FORWARDING_BACKWARDING}.
Even with cloud servers and workstations, out-of-memory is often the roadblock of machine learning tasks, which is even more critical in mobile and embedded devices.
By calculating the peak memory requirement beforehand, engineers can plan machine learning tasks more effectively and try more diverse hyper-parameters, different model structures, or higher resource utilization (e.g., increased batch sizes).

\subsection{Reduced and Proactive Swap}

\begin{figure}[t]
   \begin{center}
    \includegraphics[width=0.98\columnwidth]{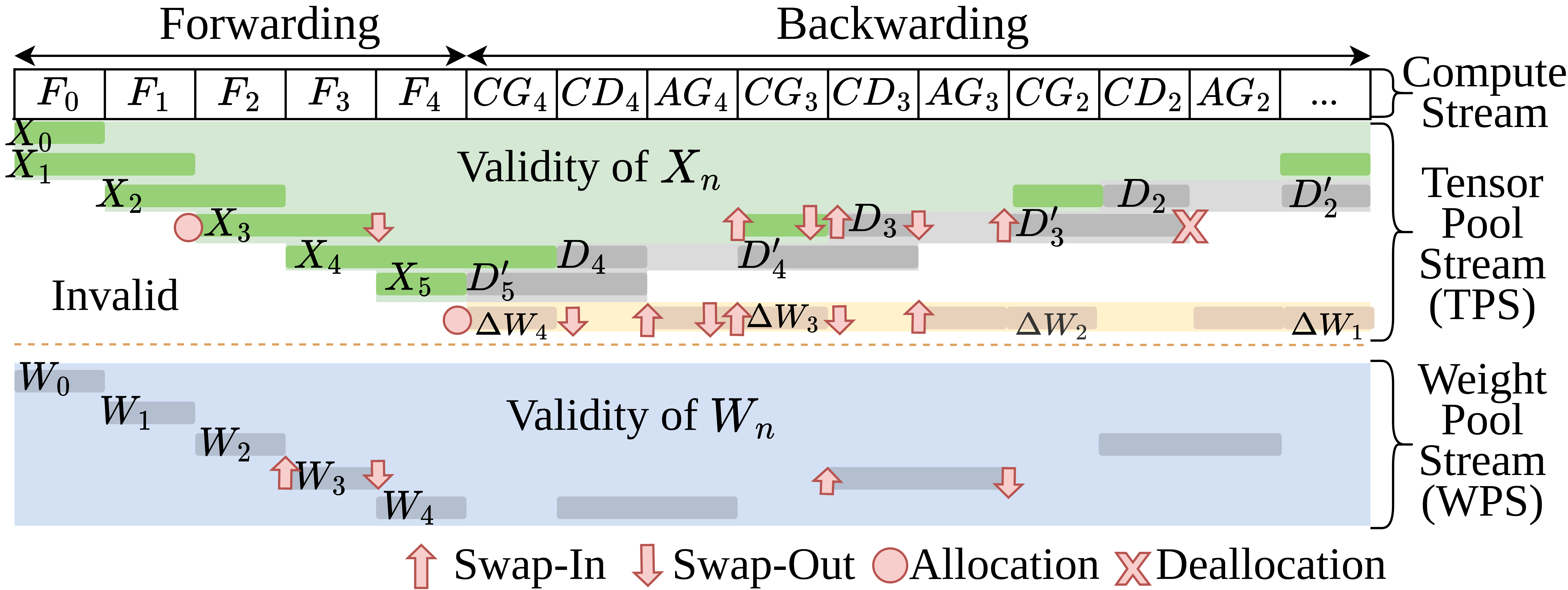}
  \end{center}
   \caption{On-demand Swap (5 linear layers).}
   \label{FIG_ONDEMAND_SWAPPING}
\end{figure}

\begin{figure}[t]
  \begin{center}
    \includegraphics[width=0.98\columnwidth]{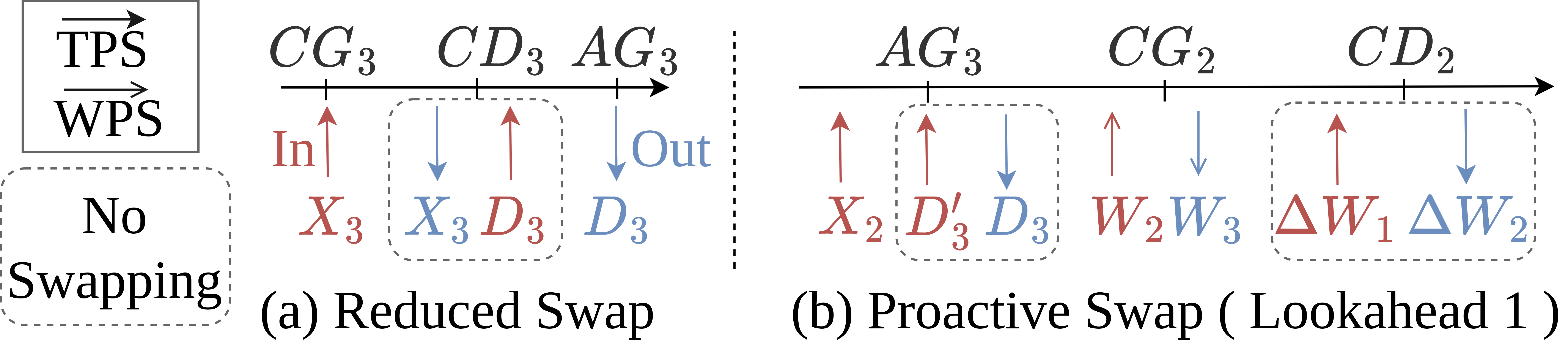}
  \end{center}
   \caption{Reduced and Proactive Swap (Lookahead 1).}
   \label{FIG_PROACTIVE_SWAPPING}
\end{figure}
Most of the previous works focus on reducing the intermediate activation which occupies a huge part of memory during training. In order to resolve this problem, \textit{NNTrainer} proposes the \textit{Proactive Swap} scheme on top of \textit{NNTrainer} memory planning based on EO.
To schedule swapping, the validity of tensors, which differs for each training procedure, should be considered.
The weight tensors should be valid throughout the whole training procedure without any allocation and deallocation (Light blue area in Figure~\ref{FIG_ONDEMAND_SWAPPING}), and the input and output tensors for activation should be valid until gradient computation whose inputs are the results of forwarding computation (Light green area in Figure~\ref{FIG_ONDEMAND_SWAPPING}).
\textit{NNTrainer} uses individual I/O streams for each Tensor Pool Stream (TPS) and Weight Pool Stream (WPS) considering different characteristics of tensors. The simplest swap scheme is On-Demand in Figure~\ref{FIG_ONDEMAND_SWAPPING} which conducts swap-in/out when required without any concern about EOs.

Figure~\ref{FIG_PROACTIVE_SWAPPING} shows an example of how Reduced and Proactive Swap works with three linear layers.
In Figure~\ref{FIG_PROACTIVE_SWAPPING}.a, the memory scheduler of \textit{NNTrainer} counts temporal and spatial relations, and reduces the number of swapping by reusing input tensors (\(X_3\)) for input derivatives (\(D_3\)) between \(CG_3\) and \(CD_3\).
Figure~\ref{FIG_PROACTIVE_SWAPPING}.b shows a Proactive Swap example with +1 lookahead from \(AG_3\) to \(CD_2\); Proactive Swap at the \textit{n}'th EO hides I/O overhead by pre-loading tensors of \textit{n+lookahead} EO and offloading tensors of \textit{n-1}'th.
For each EO, an I/O stream swaps out tensors used by the previous EO and swaps in tensors for the next (lookahead +1) EO.
We can reduce the swap overhead further by merging conflicting swap requests; e.g., \(D'_3\) and \(D_3\) at \(AG_3\).
Then, we can compute the memory overhead of Proactive Swap.
Tensors in memory at \(AG_3\) are \(X_3, D'_3, \Delta W_2\), and \(W_3\) after Proactive Swap.
Only \(W_3\) is the memory required additionally from the on-demand swap, and it is the bearable cost for the reduced latency.

The lookahead is a hyper-parameter determined statically at the initialization, and its default value is one.
However, the optimal lookahead differs by network configurations and execution orders; e.g., gradient clipping, distribution of non-trainable layers, and layer types.
Thus, we can further reduce memory and computation overhead by assigning lookahead for each EO dynamically; i.e., let each lookahead value converge throughout iterations.
This is left for future work.

\section{Evaluation}\label{S:Evaluation}

We evaluate both small experimental neural networks and large practical neural networks.
We analyze and compare latency and memory consumption of \textit{NNTrainer} and conventional frameworks: PyTorch 1.13.1, TensorFlow 2.11.0, TensorFlow-Lite 2.11.0 (both C++ and Python implementations); recent TensorFlow-Lite supports training.
%In order to analyze the performance of \textit{NNTrainer} in more detail, we first evaluate the components of the neural network, Layers and discuss about AI Applications including commonly used neural network models.
For memory consumption, we also compare with the theoretical memory requirement based on Analysis (\S\ref{S:Analysis}).
We demonstrate on-device training applications on devices: a complex text-to-speech (TTS) application with multiple LSTM layers on Galaxy S21 Ultra (Exynos 2100, 1\(\times\)ARM Cortex-X1 2.9 GHz, 3\(\times\)ARM Cortex-A78 2.9GHz, 4\(\times\)Cortex-A55 2.2GHz) and a Transformer model with huge memory requirements, virtually impossible for embedded devices to run.
Except for TTS, we evaluate with Raspberry pi 4, which has 4\(\times\)ARM Cortex-A72 (1.5GHz) CPU cores and 8 GiB RAM running Ubuntu 22.04.
We evaluate \textit{NNTrainer} with CPUs as its computation backend.
Note that most NPUs and DSPs of embedded devices do not support floating points and GPUs of consumer electronics are often not available for training; e.g., GPUs of TV are supposed to be fully dedicated to video streams and GUIs.

\subsection{Component Evaluation}

Table~\ref{TBL_COMPONENT_EVAL} describes 5 test cases of small neural networks with different dimensions of inputs and labels.
The test cases include an LSTM layer (popular for time series including voice models) and linear and Conv2D layers (popular for vision models).
We evaluate a case of three linear layers (FC-FC-FC) with non-trainable layers in the middle and a case of Figure~\ref{FIG_EXE_ORDER_FL} with Conv2D layers (Conv-AC-FL). 
We have applied mean squared error (MSE)~\cite{allen1971mean} and stochastic gradient descent (SGD)~\cite{ketkar2017stochastic} to train the test cases.

We ensure the correctness of \textit{NNTrainer} by comparing every activation and weight value of models trained by \textit{NNTrainer} with the same models and data trained by TensorFlow.
The two frameworks result in equivalent neural network models with errors at \(10^{-4}\) level.
Note that an automated test suite~\cite{lim2021lightsys} ensures the correctness and equivalence for each pull request in different environments.
%With automated procedures, we confirm the correctness by training models and examining their equivalence for every pull request; i.e., if a weight or activation value has an error over \(1e-4\), the corresponding code commit is automatically rejected.

\begin{table}[tb]
  \centering
  \begin{threeparttable}
    \resizebox{\columnwidth}{!}{
    \begin{tabular}{c c c c}
    Test Cases & Input & Output (Label) & Theoretical Memory (KiB)\\
     \hline 
      Linear & 64:1:1:150528& 64:1:1:300 & \(390590\)\\
      Conv2D & 64:3:224:224& 64:3:112:112& \(65856\)\\
      LSTM &64:1:1:150528& 64:1:1:10&\(84731\)  \\
      FC-FC-FC & 2048:1:1:784& 2048:1:1:100 &\(135928\) \\
      Conv-AC-FL & 64:3:224:224& 64:1:1:37632 & \(65856\)\\ 
    \hline
    \end{tabular}}
  \end{threeparttable}
  \caption{Test cases for component evaluation.}
  \label{TBL_COMPONENT_EVAL}
\end{table}

As explained in Section~\ref{S:Analysis}, once the input sizes and configurations are determined, the theoretical memory requirements can be calculated based on the sizes of intermediate activations, weights, and gradients as in Table~\ref{TBL_COMPONENT_EVAL}.

\begin{figure}[t]
  \begin{center}
    \includegraphics[width=0.98\columnwidth]{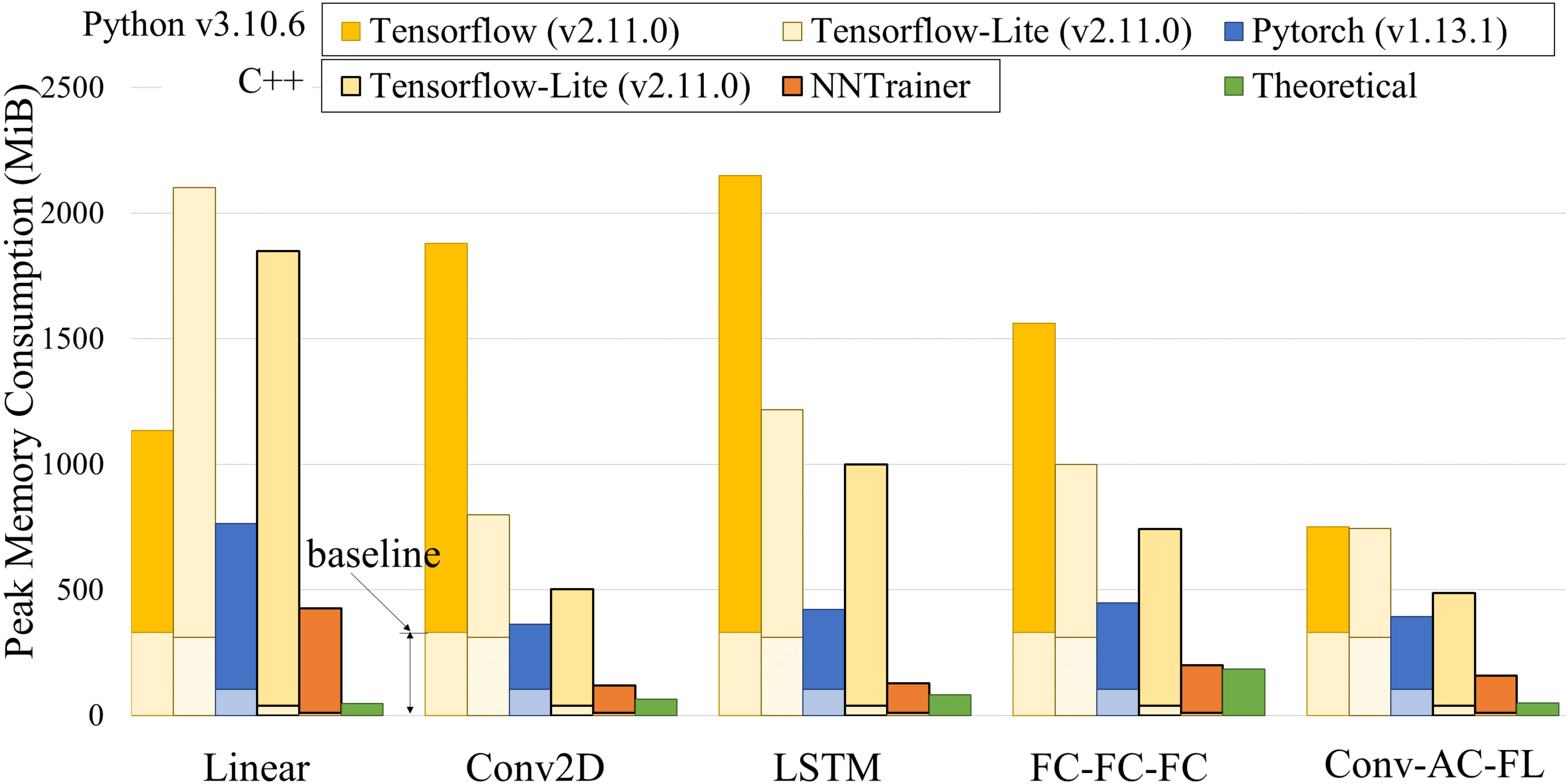}
  \end{center}
  \caption{Peak memory consumption}
  \label{FIG_COMPONENT_MEM_EVAL}
\end{figure}

\begin{figure}[t]
  \begin{center}
    \includegraphics[width=0.98\columnwidth]{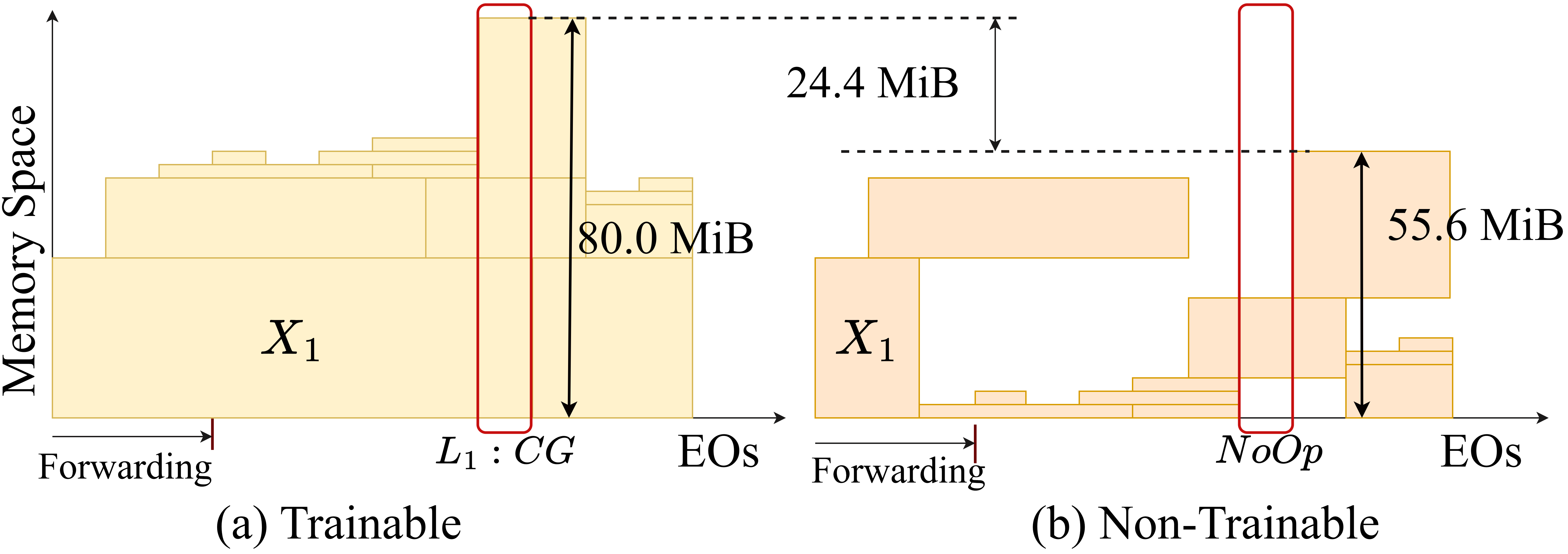}
  \end{center}
  \caption{Non-trainable layer analysis}
  \label{FIG_NON_TRAINABLE_ANAL}
\end{figure}

Figure~\ref{FIG_COMPONENT_MEM_EVAL} shows the evaluation results of the baseline and peak memory consumption.
The baseline implies memory consumed by the framework itself (e.g., codes and libraries), which is constant for different test cases.
The peak implies the memory consumed by training a model in addition to the baseline.
TensorFlow consumes x27 of baseline memory (329.8 MiB) than \textit{NNTrainer} (11.3 MiB) consumes, and PyTorch consumes x8 (103.4 MiB).
TensorFlow-Lite in C++ (39.6 MiB) consumes three times.
Such excessive baseline memory consumption of TensorFlow, TensorFlow-Lite(python) and PyTorch can be attributed to Python and a lot of external libraries; \textit{NNTrainer} and TensorFlow-Lite (C++) are written in C++ with minimal dependencies on external libraries.
The peak memory consumption varies per test case, and, unlike the baseline, the difference is mostly caused by design choices (including granularity of training procedures) and buffer management approaches affected by the choices, not by the libraries and languages for the implementation.

In single and multi-layer tests of Linear, Conv2D, and LSTM, \textit{NNTrainer} consumes significantly less amount of memory.
The proposed resource utilization method based on EOs allows reusing memory spaces for Tensors between layers.
Note that conventional frameworks consume significantly larger memory than \textit{NNTrainer} does: x2.52 to x11.47 on average including baseline.
Even Tensorflow-Lite in C++ consumes x4 of that \textit{NNtrainer} consumes.
The theoretical memory requirement in Figure~\ref{FIG_COMPONENT_MEM_EVAL} suggests that \textit{NNTrainer} is extremely efficient in using memory by sharing the most shareable tensors with ignorable memory overhead.
The baseline is required to load essential libraries, and additional heaps attributing to the peak are required for some layers; e.g., 
\textit{NNTrainer}'s Conv2D layer adds ``Image to Column'' (im2col)~\cite{dukhan2019indirect} operator for computation efficiency, which requires additional buffers.

We evaluate how much memory can be reduced by \textit{NNTrainer} with FC-FC-FC case by setting the second layer non-trainable.
Figure~\ref{FIG_NON_TRAINABLE_ANAL} shows the memory placement along with EOs by \textit{NNTrainer}'s memory planner. 
\(X_1\) is the input tensor of the second layer which needs to be saved for computing gradient (\(L_1:CG)\) for the training case (Figure~\ref{FIG_NON_TRAINABLE_ANAL}.a).
For the non-trainable case, the validity of \(X_1\) is during forwarding as in Figure~\ref{FIG_NON_TRAINABLE_ANAL}.b so that the memory space is utilized by other tensors.
The results show the memory reduction at \(L_1:CG\) (red rectangle), the non-trainable layer.
However, the peak memory is measured at a different EO and 24.4 MiB is expected to be reduced.
The experimental results show that 20.9 MiB is reduced: 146.1 MiB vs. 125.2 MiB.

\begin{figure}[t]
  \begin{center}
    \includegraphics[width=0.98\columnwidth]{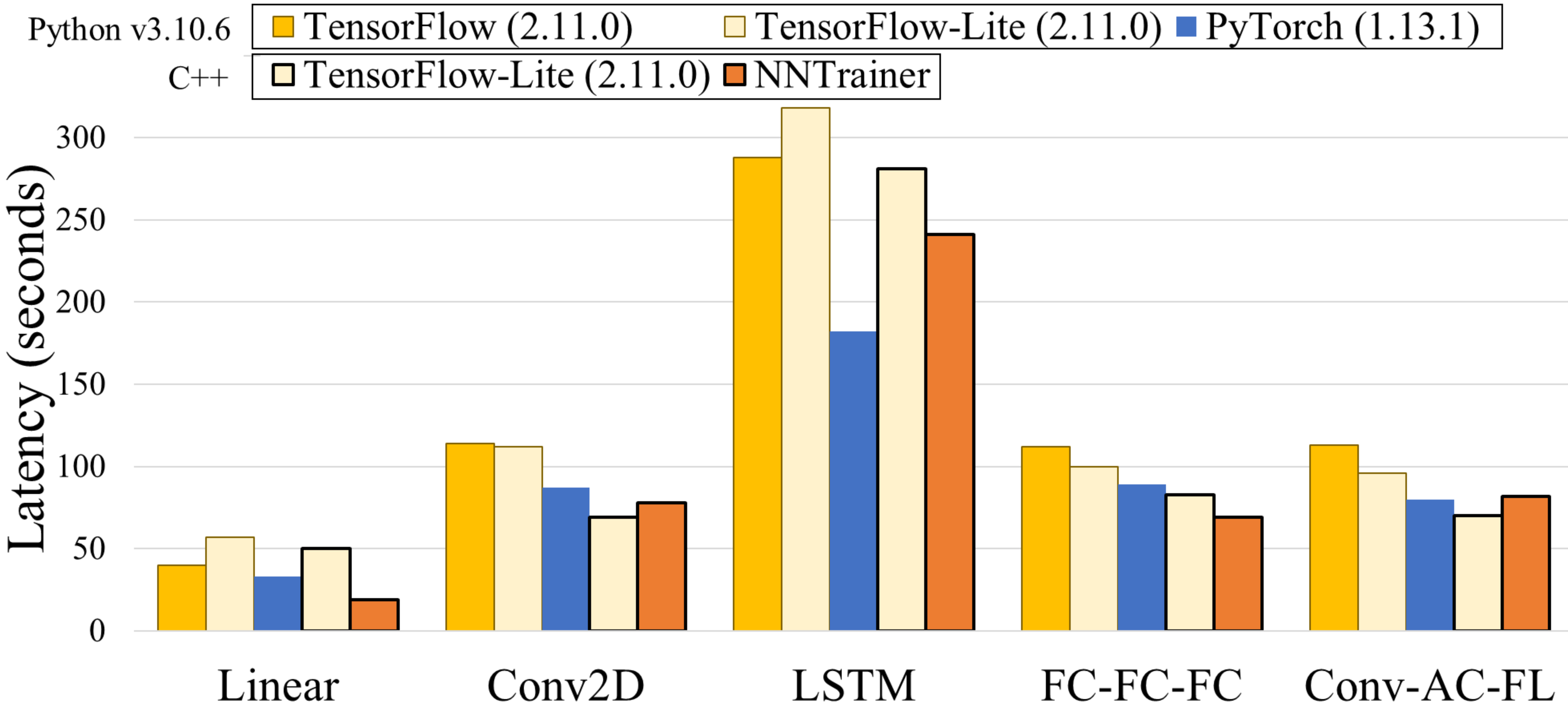}
  \end{center}
  \caption{Training latency of the test cases from Table~\ref{TBL_COMPONENT_EVAL}.}
  \label{FIG_COMPONENT_SPEED_EVAL}
\end{figure}
\begin{figure*}[ht]
  \centering\includegraphics[width=1.0\textwidth]{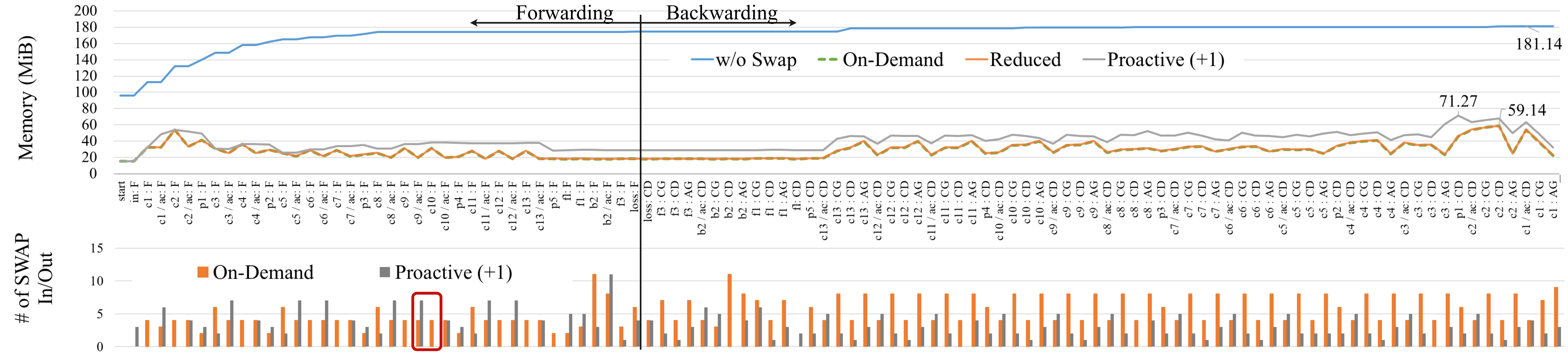}
  \caption{VGG16 training performance of swap schemes. On-Demand and Reduced show the exactly same performance.}
  \label{FIG_SWAP_RESULTS}
\end{figure*}

Figure~\ref{FIG_COMPONENT_SPEED_EVAL} shows the training latency of 1 epoch, 640 dataset size with the test cases in Table~\ref{TBL_COMPONENT_EVAL}.
This experimental result shows that \textit{NNTrainer} does not sacrifice the latency or the accuracy to conserve memory.
Although consuming significantly less memory, in most cases, \textit{NNTrainer} is evaluated to be faster than or equivalent to the conventional frameworks.

\subsection{On-Demand, Reduced, Proactive Swap}

\begin{figure}[t]
  \begin{center}
    \includegraphics[width=0.98\columnwidth]{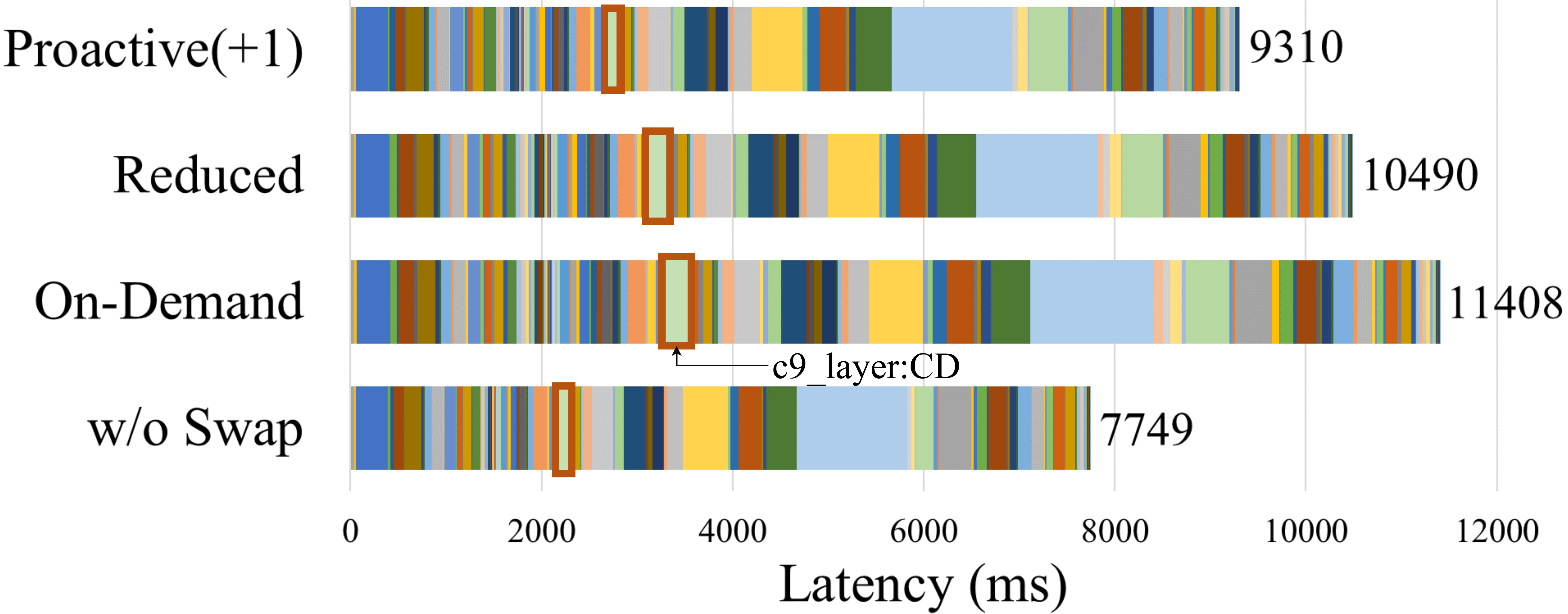}
  \end{center}
  \caption{Latency of Proactive Swap (VGG16)}
  \label{FIG_SWAP_LATENCY}
\end{figure}

Swapping allows additional memory saving selectively; users may turn it on and off.
We compare the peak memory consumption and the number of swapping for different schemes, On-Demand, Reduced, and Proactive Swap with one lookahead parameter, in Figure~\ref{FIG_SWAP_RESULTS}.
Proactive Swap includes Reduced Swap.
We train the VGG16 model with 64 batches of \(32\times32\times3\) inputs.
Figure~\ref{FIG_SWAP_LATENCY} shows the latency of each EO for each scheme.
As expected, without swapping, the largest size of memory is consumed, although it is still the smallest among the compared frameworks in Figure~\ref{FIG_PEAK_MEMORY_CONSUMPTION}.
The evaluations show that Proactive Swap reduces the peak memory consumption from 181 MiB to 71 MiB successfully with 20.1\% latency overhead and On-Demand Swap reduces to 51 MiB with 35.5\% latency overhead.
The peak memory consumption of On-Demand Swap is 59 MiB, the same as Reduced Swap at c2:CD.
The memory size of input derivative of c2:CD, \(\Delta D'\) is \(64\times32\times32\times64\times\)\texttt{\small{sizeof(float)}} and output derivative \(\Delta D\) has the same size. The memory size for weight \(W\) is \((64\times3\times3\times64 + 64)\times\)\texttt{\small{sizeof(float)}}.
Then, the theoretical memory size, which needs to be kept allocated at c2:CD, is 32.14 MiB.
Considering the baseline (11 MiB), this is close to the peak memory consumption with swapping.
Thus, we can conjecture that \textit{NNTrainer}'s swap scheduling utilizes memory extremely efficiently.

There is almost the same memory reduction achieved for both On-Demand and Reduced Swap; however, Reduced Swap reduces the latency by 10\% with reduced number of swapping by the EO-based memory planning proposed in \S\ref{S:Design}. 
Proactive Swap consumes more memory while reducing latency by 10\%; it further reduces swapping as shown in Figure~\ref{FIG_PROACTIVE_SWAPPING}.b and Figure~\ref{FIG_SWAP_RESULTS}.
Note that Proactive Swap does not request swapping at a convolution layer during forwarding (a red box in Figure~\ref{FIG_SWAP_RESULTS}).
It is because its activation layer incurs in-place computation, which does not need additional memory swap-in.
However, we need a swap-in after the activation before its next layer; thus, the swapping count increases. 
The latency of c9\_layer:CD (green boxes with red borders) in Figure~\ref{FIG_SWAP_LATENCY} compares the latency of the swap schemes.

\subsection{Applications}
\begin{figure}[t]
  \begin{center}
    \includegraphics[width=0.98\columnwidth]{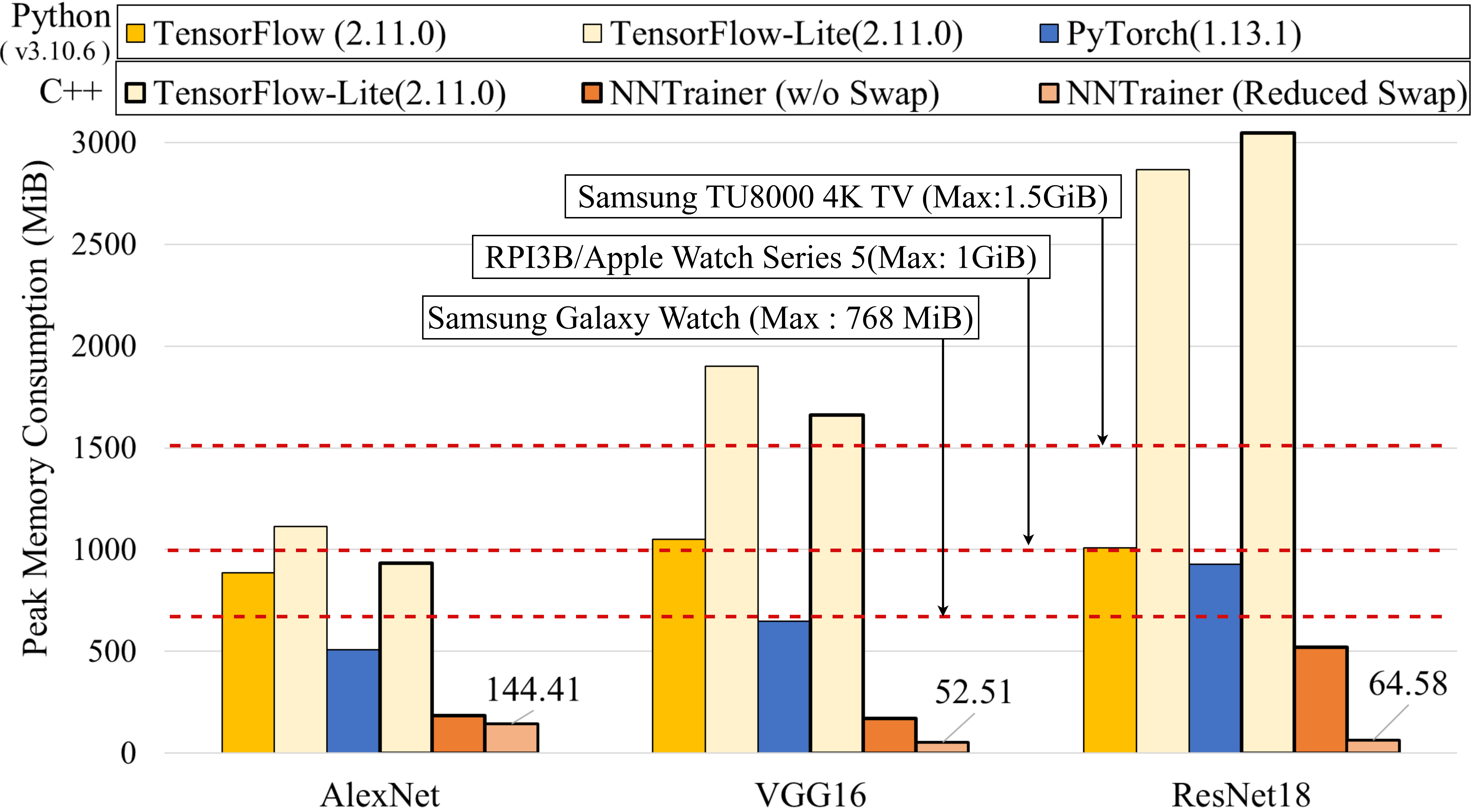}
  \end{center}
  \caption{Memory consumption of training applications.}
  %MNIST (LeNet-6), CiFar-100~\cite{krizhevsky2009learning} (VGG16, ResNet18, and Pretrained Model for Transfer Learning), CiFar-10 (Target Datset of Transfer Learning), and Movielens (product rating).}
  \label{FIG_PEAK_MEMORY_CONSUMPTION}
\end{figure}

Figure~\ref{FIG_PEAK_MEMORY_CONSUMPTION} shows the peak memory consumption of training with the batch size of 64, trained by \textit{NNTrainer} (w/o Swap, Reduced Swap), TensorFlow, TensorFlow-Lite (Python, C++), and PyTorch.
The models are trained from scratch: AlexNet~\cite{krizhevsky2017imagenet}, VGG16~\cite{simonyan2014very}, and Resnet18~\cite{he2016deep}.
In every case, \textit{NNTrainer} consumes much less memory.
Without swap, it consumes 23\% (AlexNet), 15\% (VGG16), and 35\% (ResNet18) of what the other uses.
With Reduced Swap, it consumes 18\% (AlexNet) and 5\% (VGG16 and ResNet18) of what the other uses; the most efficient of others, PyTorch, consumes 20 times more!
AlexNet shows less improvement with swapping is because the last two linear layers consume peak memory; the weight dimension is \(4096\times4096\), which is 62 MiB.
The theoretical peak memory of Apply Gradient is two times of its weight size due to the gradient (124 MiB).
Tensorflow-Lite, with both C++ and Python, consumes almost x46 times more memory for ResNet18 even if it is for embedded devices.
\begin{figure}[t]
  \begin{center}
     \includegraphics[width=0.98\columnwidth]{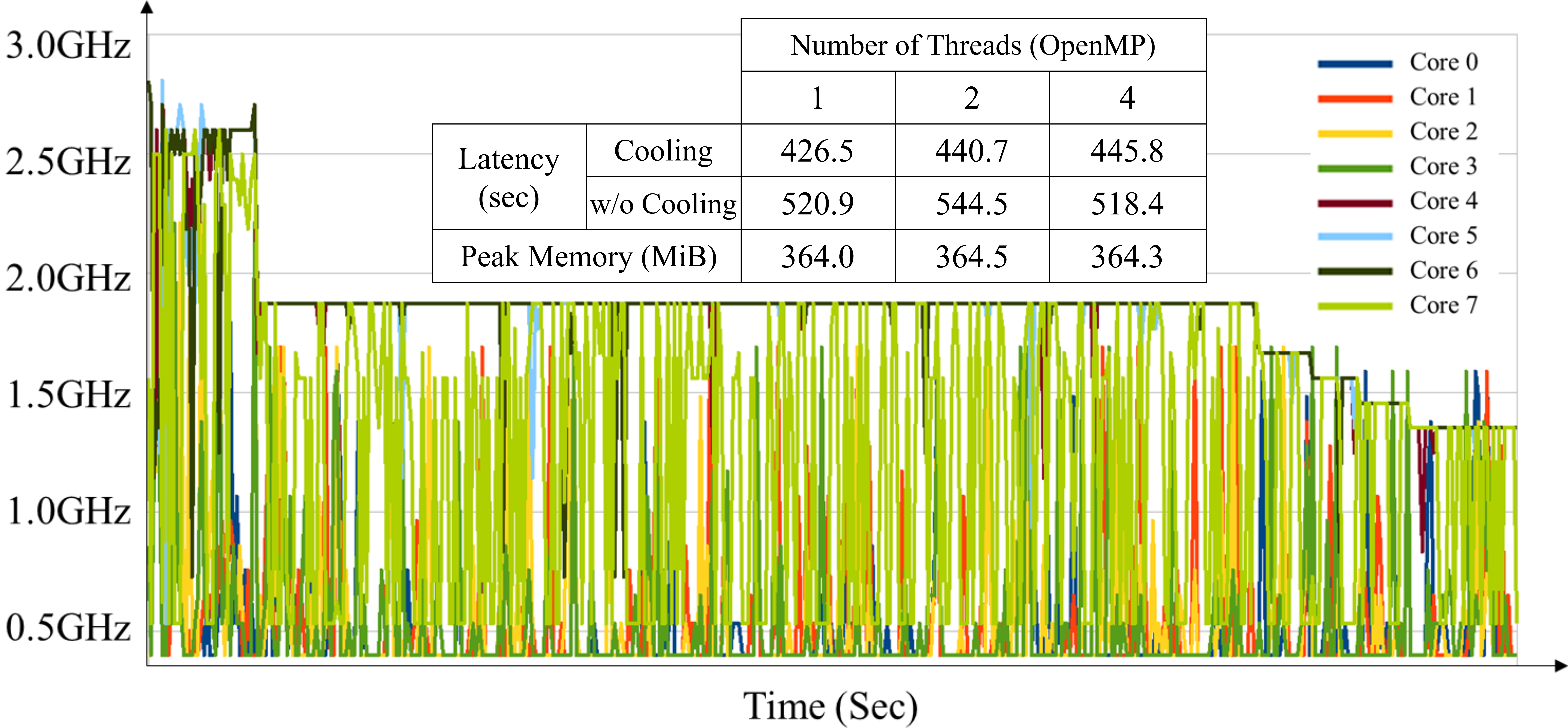}
   \end{center}
   \caption{CPU throttling of Tacotron2 (Galaxy S21 Ultra)}
   \label{FIG_TACOTRON2}
\end{figure}

\textbf{Tacotron2}~\cite{shen2018natural} is a sequence-to-sequence text-to-speech model based on attention, which enables end-to-end training with dataset of sentence and voice.
Developers can create a personalized text-to-speech (TTS) application by running Tacotron2 on \textit{NNTrainer} so that an application may read books for kids with voices of their parents or translate with the user's voice.
%The encoder plays a role in converting a character-unit one-hot vector into an encoder features, and attention part extracts the features to be used in the decoder among them.
%Lastly, decoder uses this feature and the mel-spectrogram generated at the previous time iteration to compute for the next time iteration.
%The purpose of personalization of tacotron2 in this paper is to use the user's voice when the text is converted to speech.
In our application, a user reads 18 sentences to create a user dataset for personalization, which is combined with the pre-installed dataset.
Tacotron2 consists of an Encoder, Attention, and Decoder, and the Decoder generates voice outputs; thus, we apply on-device training to the Decoder, and other parts are kept static for inferences.
%%%%%%%% MJ: TACOTRON를 우리가 이용하지 않는 부분까지 더 자세하게 묘사할 필요는 없어 보입니다. 혹시 규모를 묘사하려면 총 레이어 갯수나 파라메터 갯수가 낫습니다.
%Decoder consists of Prenet, Decoder LSTMs, and Postnet.
%Prenet has 2 linear layers.
%Decoder LSTMs has 2 LSTM layers including attention and 2 linear layers for gate prediction and a mel spectrogram.
%Postnet has 5 Conv1D layers, and it runs after time iteration is finished by concatenating the spectrogram computed with time iterations.

\textit{NNTrainer} does time-unrolling to perform complex time iteration, and the entire memory is statically allocated based on the maximum time iteration declared by developers.
Weights of the same layers that are time-unrolled incur no additional memory or computation because of the Tensor sharing.
Because of time iteration, forward process is performed for all unrolled layers, and gradients are accumulated in the backward process without updating weights.
The optimizer updates weights only once per layer, which requires additional memory to store the gradient during time iterations. Gradient Clipping~\cite{pascanu2013difficulty} and Teacher Forcing~\cite{williams1989learning} are also supported.

In Figure~\ref{FIG_TACOTRON2}, peak memory consumption and latency with the different numbers of threads for training with 26 samples on Galaxy S21 Ultra are measured to evaluate the effect of parallelism with OpenBLAS~\cite{blas}. 
Overheating is a major issue of mobile phones, and to mitigate it, CPU cores are usually throttled by limiting their frequencies.
Training neural networks heavily utilize CPUs; thus, we observe how the CPU throttling affects the latency. Figure~\ref{FIG_TACOTRON2} shows the latency and frequency of each core, and we can see that the CPU frequencies are continuously throttled down.
Another finding is that multi-core parallelism is not helpful for this model, and if we can cool CPUs properly, the training performance will be improved significantly.
Only 364 MiB of memory is requested for the case without swap to personalize the Tacotron2 and \textit{NNTrainer} completes within ten minutes even with throttled CPU cores, which has been enough to pass the quality assurance team.

\begin{figure}[t]
  \begin{center}
     \includegraphics[width=0.98\columnwidth]{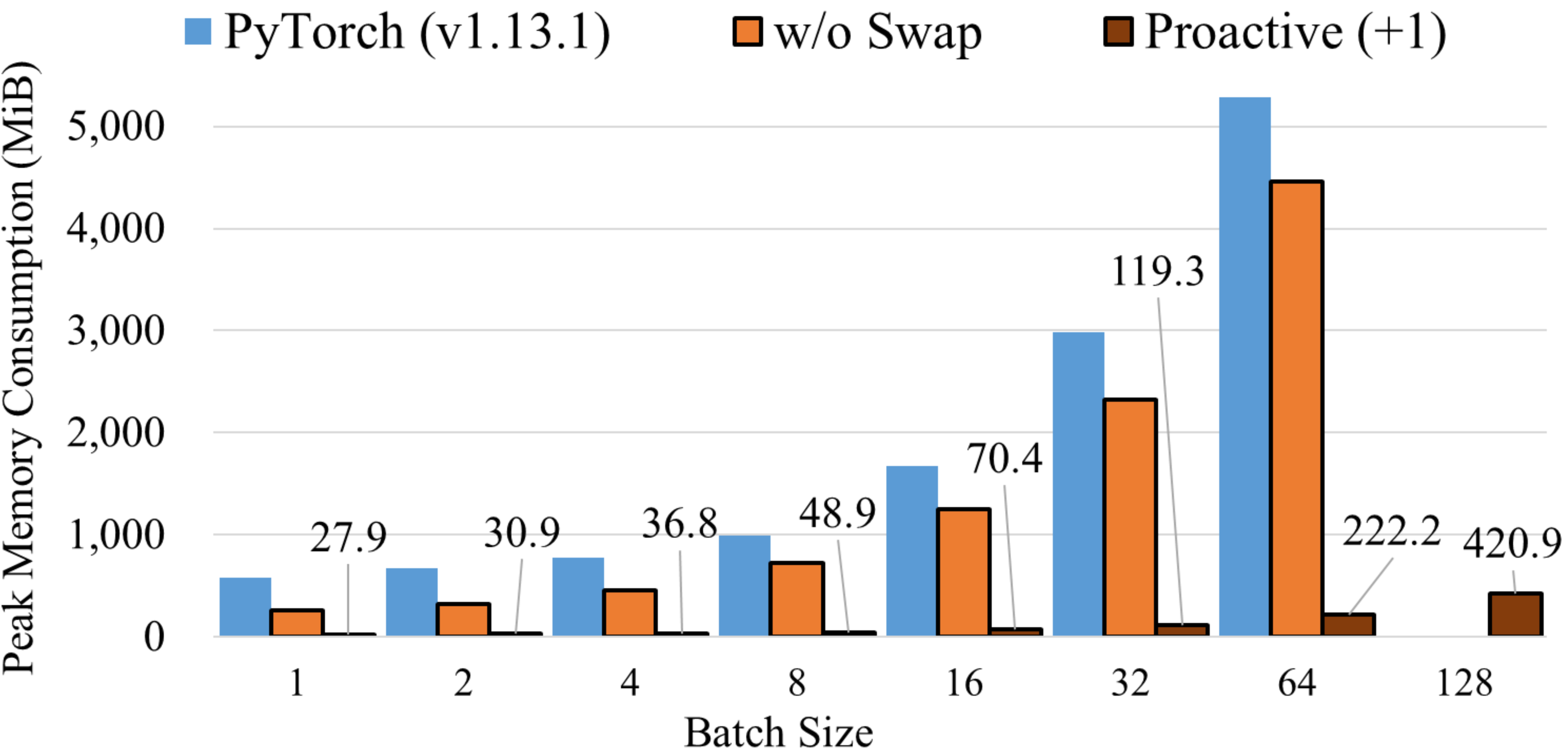}
   \end{center}
   \caption{Peak Memory Consumption of Transformer}
   \label{FIG_TRANSFORMER}
\end{figure}

\textbf{Transformer}~\cite{vaswani2017attention} is a popular neural network model throughout various applications.
Transformer consists of Encoder and Decoder, and each of them has stacked multi-head attention layers, which require a huge amount of memory: over 6 GiB for 128 batch size, 6 stacked 8 multi-headed attention layers in both encoder and decoder to train.
Although it requires memory-intensive computation, each multi-head attention layer is independent, promoting easier and more efficient parallelism.
Conventional frameworks intentionally allocate weights of \(Q, K, V\) for each head and try to achieve shorter latency with parallelism.
It is definitely efficient in clouds, which have abundant resources; however, it is usually not desirable on devices because the size of the internal activation of the attention layer increases.
For swapping, the smaller the size of the layer intermediate activation or weight, the more reduction in the memory consumption.
Therefore, \textit{NNTrainer} implements the multi-head attention using \(Concat(Attention(QV_i^Q, KW_i^K, VW_i^V))_h\) as proposed in \cite{vaswani2017attention}.
Figure~\ref{FIG_TRANSFORMER} shows the size of memory consumed to train Transformer with different batch sizes.
With relatively small batch sizes, the effect of memory planner based on EOs is noticeable compared to PyTorch; it is almost half (575.2 vs 256.9 MiB).
However, as the batch size increases, the size of intermediate activation saved for gradient computing grows exponentially and the effect of the memory planer diminishes.
Proactive Swap reduces the memory consumption dramatically from 4.4 GiB to 0.22 GiB (down to 1/20) for 64 batch size.
Moreover, only Proactive Swap can train with 128 batch size in a machine with 8 GiB RAM, utilizing only 420 MiB.

\section{Future Works}\label{S:Future_works}
Our future plan is to make \textit{NNTrianer} more widely available and expand for wider range of machine learning techniques popular in embedded device applications. Also, we do have a plan to develop new techniques to utilize resources more efficiently.
\begin{itemize}
\item Extend the reach of \textit{NNTrainer}. This includes providing an interface with other frameworks, where model importers will be our first priority followed by model exporters, and expand platform support for Android with Java APIs, support Yocto/OpenEmbedded and TIZEN with C\# and Web APIs.
\item Extend support to various computation backends and accelerators for embedded devices such as GPUs, DSPs and NPUs to expand computational capacity with higher energy efficiency. Integer-based training or quantization-aware training~\cite{jacob2018quantization} is the next step.
\item Support advanced few-shot learning techniques such as meta-learning and adaption problems.
\item Support half precision during training. This is more efficient on-device at a point of latency and memory consumption.
%\item Develop efficient resource utilization techniques. By enabling the control ability of the training process to the Graph module, it is possible for developer to manage more freely and develop their application efficiently. Although \textit{NNTrainer} uses less memory, \textit{Dynamic off-loading} using second storage is help to save more memory. In general, since training takes a lot of time and the layer-by-layer based operation is typical, most of the data loaded in memory is not in an active state except for the data of the layer participating in the computation. Dynamic off-loading is a technique that loads only the necessary memory in the layer currently participating int the calculation. The \textit{NNTrainer} decides when to load which data considering the computation and  the data transfer latency by itself. Once \textit{NNTrainer} supports it, we can train with much less memory.
\end{itemize}

\section{Conclusions}\label{S:Conclusions}

We propose a highly efficient and light-weight on-device neural network training framework, \textit{NNTrainer}, with techniques based on novel observations on neural network training mechanisms.
We significantly reduce memory consumption so that embedded devices can practically train neural networks without deterioration of accuracy or modification of model architecture.
The evaluation results show the efficiency of \textit{NNTrainer} for various neural network models and applications.
The proposed framework, \textit{NNTrainer}, is practical open-source software that personalizes complex AI services on mass-produced devices.

%\begin{acks}
%This project is funded and sponsored by Samsung Research of Samsung Electronics.
%Linux Foundation AI \& Data is another sponsor that supports DevOps and the open-source community of this project.
%We appreciate developers in business divisions or other research teams who have participated in use case refinement, architecture design, and implementations.
%%%%%%%%%%% 사업부 개발자 중 Major한 분들???
%We also appreciate every open-source contributor who has raised issues, suggested features, and proposed code commits; major contributors who have dedicated to the development include Parichay Kapoor, Hyeonseok Lee, Jihoon Lee, Dongju Chae, Jiho Chu, Youngjae Shin, and Chani Lee.
%\end{acks}

\bibliographystyle{ACM-Reference-Format}
\bibliography{main}


\begin{thebibliography}{50}

%%% the Web address.

\ifx \showCODEN    \undefined \def \showCODEN     #1{\unskip}     \fi
\ifx \showDOI      \undefined \def \showDOI       #1{#1}\fi
\ifx \showISBNx    \undefined \def \showISBNx     #1{\unskip}     \fi
\ifx \showISBNxiii \undefined \def \showISBNxiii  #1{\unskip}     \fi
\ifx \showISSN     \undefined \def \showISSN      #1{\unskip}     \fi
\ifx \showLCCN     \undefined \def \showLCCN      #1{\unskip}     \fi
\ifx \shownote     \undefined \def \shownote      #1{#1}          \fi
\ifx \showarticletitle \undefined \def \showarticletitle #1{#1}   \fi
\ifx \showURL      \undefined \def \showURL       {\relax}        \fi
% invisible to TeX
\providecommand\bibfield[2]{#2}
\providecommand\bibinfo[2]{#2}
\providecommand\natexlab[1]{#1}
\providecommand\showeprint[2][]{arXiv:#2}

\bibitem[Abadi et~al\mbox{.}(2016)]%
        {199317}
\bibfield{author}{\bibinfo{person}{Mart{\'\i}n Abadi}, \bibinfo{person}{Paul Barham}, \bibinfo{person}{Jianmin Chen}, \bibinfo{person}{Zhifeng Chen}, \bibinfo{person}{Andy Davis}, \bibinfo{person}{Jeffrey Dean}, \bibinfo{person}{Matthieu Devin}, \bibinfo{person}{Sanjay Ghemawat}, \bibinfo{person}{Geoffrey Irving}, \bibinfo{person}{Michael Isard}, \bibinfo{person}{Manjunath Kudlur}, \bibinfo{person}{Josh Levenberg}, \bibinfo{person}{Rajat Monga}, \bibinfo{person}{Sherry Moore}, \bibinfo{person}{Derek~G. Murray}, \bibinfo{person}{Benoit Steiner}, \bibinfo{person}{Paul Tucker}, \bibinfo{person}{Vijay Vasudevan}, \bibinfo{person}{Pete Warden}, \bibinfo{person}{Martin Wicke}, \bibinfo{person}{Yuan Yu}, {and} \bibinfo{person}{Xiaoqiang Zheng}.} \bibinfo{year}{2016}\natexlab{}.
\newblock \showarticletitle{TensorFlow: A System for Large-Scale Machine Learning}. In \bibinfo{booktitle}{\emph{12th {USENIX} Symposium on Operating Systems Design and Implementation ({OSDI} 16)}}. \bibinfo{publisher}{{USENIX} Association}, \bibinfo{address}{Savannah, GA}, \bibinfo{pages}{265--283}.
\newblock
\showISBNx{978-1-931971-33-1}
\urldef\tempurl%
\url{https://www.usenix.org/conference/osdi16/technical-sessions/presentation/abadi}
\showURL{%
\tempurl}


\bibitem[Allen(1971)]%
        {allen1971mean}
\bibfield{author}{\bibinfo{person}{David~M Allen}.} \bibinfo{year}{1971}\natexlab{}.
\newblock \showarticletitle{Mean square error of prediction as a criterion for selecting variables}.
\newblock \bibinfo{journal}{\emph{Technometrics}} \bibinfo{volume}{13}, \bibinfo{number}{3} (\bibinfo{year}{1971}), \bibinfo{pages}{469--475}.
\newblock


\bibitem[Baydin et~al\mbox{.}(2018)]%
        {baydin2018automatic}
\bibfield{author}{\bibinfo{person}{Atilim~Gunes Baydin}, \bibinfo{person}{Barak~A Pearlmutter}, \bibinfo{person}{Alexey~Andreyevich Radul}, {and} \bibinfo{person}{Jeffrey~Mark Siskind}.} \bibinfo{year}{2018}\natexlab{}.
\newblock \showarticletitle{Automatic differentiation in machine learning: a survey}.
\newblock \bibinfo{journal}{\emph{Journal of machine learning research}}  \bibinfo{volume}{18} (\bibinfo{year}{2018}).
\newblock


\bibitem[Bulatov(2018)]%
        {gradientchecking}
\bibfield{author}{\bibinfo{person}{Yaroslav Bulatov}.} \bibinfo{year}{2018}\natexlab{}.
\newblock \bibinfo{title}{Fitting larger networks into memory}.
\newblock \bibinfo{howpublished}{\url{https://medium.com/tensorflow/fitting-larger-networks-into-memory-583e3c758ff9}}.
\newblock


\bibitem[Cai et~al\mbox{.}(2020)]%
        {cai2020tinytl}
\bibfield{author}{\bibinfo{person}{Han Cai}, \bibinfo{person}{Chuang Gan}, \bibinfo{person}{Ligeng Zhu}, {and} \bibinfo{person}{Song Han}.} \bibinfo{year}{2020}\natexlab{}.
\newblock \showarticletitle{Tinytl: Reduce memory, not parameters for efficient on-device learning}.
\newblock \bibinfo{journal}{\emph{arXiv preprint arXiv:2007.11622}} (\bibinfo{year}{2020}).
\newblock


\bibitem[Chen et~al\mbox{.}(2016)]%
        {chen2016training}
\bibfield{author}{\bibinfo{person}{Tianqi Chen}, \bibinfo{person}{Bing Xu}, \bibinfo{person}{Chiyuan Zhang}, {and} \bibinfo{person}{Carlos Guestrin}.} \bibinfo{year}{2016}\natexlab{}.
\newblock \showarticletitle{Training deep nets with sublinear memory cost}.
\newblock \bibinfo{journal}{\emph{arXiv preprint arXiv:1604.06174}} (\bibinfo{year}{2016}).
\newblock


\bibitem[Dahl et~al\mbox{.}(2011)]%
        {dahl2011context}
\bibfield{author}{\bibinfo{person}{George~E Dahl}, \bibinfo{person}{Dong Yu}, \bibinfo{person}{Li Deng}, {and} \bibinfo{person}{Alex Acero}.} \bibinfo{year}{2011}\natexlab{}.
\newblock \showarticletitle{Context-dependent pre-trained deep neural networks for large-vocabulary speech recognition}.
\newblock \bibinfo{journal}{\emph{IEEE Transactions on audio, speech, and language processing}} \bibinfo{volume}{20}, \bibinfo{number}{1} (\bibinfo{year}{2011}), \bibinfo{pages}{30--42}.
\newblock


\bibitem[Delange et~al\mbox{.}(2021)]%
        {delange2021continual}
\bibfield{author}{\bibinfo{person}{Matthias Delange}, \bibinfo{person}{Rahaf Aljundi}, \bibinfo{person}{Marc Masana}, \bibinfo{person}{Sarah Parisot}, \bibinfo{person}{Xu Jia}, \bibinfo{person}{Ales Leonardis}, \bibinfo{person}{Greg Slabaugh}, {and} \bibinfo{person}{Tinne Tuytelaars}.} \bibinfo{year}{2021}\natexlab{}.
\newblock \showarticletitle{A continual learning survey: Defying forgetting in classification tasks}.
\newblock \bibinfo{journal}{\emph{IEEE Transactions on Pattern Analysis and Machine Intelligence}} (\bibinfo{year}{2021}).
\newblock


\bibitem[Dukhan(2019)]%
        {dukhan2019indirect}
\bibfield{author}{\bibinfo{person}{Marat Dukhan}.} \bibinfo{year}{2019}\natexlab{}.
\newblock \showarticletitle{The indirect convolution algorithm}.
\newblock \bibinfo{journal}{\emph{arXiv preprint arXiv:1907.02129}} (\bibinfo{year}{2019}).
\newblock


\bibitem[Essalmi et~al\mbox{.}(2010)]%
        {essalmi2010fully}
\bibfield{author}{\bibinfo{person}{Fathi Essalmi}, \bibinfo{person}{Leila Jemni~Ben Ayed}, \bibinfo{person}{Mohamed Jemni}, \bibinfo{person}{Sabine Graf}, {et~al\mbox{.}}} \bibinfo{year}{2010}\natexlab{}.
\newblock \showarticletitle{A fully personalization strategy of E-learning scenarios}.
\newblock \bibinfo{journal}{\emph{Computers in Human Behavior}} \bibinfo{volume}{26}, \bibinfo{number}{4} (\bibinfo{year}{2010}), \bibinfo{pages}{581--591}.
\newblock


\bibitem[Google(2018)]%
        {tflite}
\bibfield{author}{\bibinfo{person}{Google}.} \bibinfo{year}{2018}\natexlab{}.
\newblock \bibinfo{title}{TensorFlow-Lite}.
\newblock \bibinfo{howpublished}{\url{https://www.tensorflow.org/lite}}.
\newblock


\bibitem[Ham et~al\mbox{.}(2021)]%
        {ham2021nnstreamer}
\bibfield{author}{\bibinfo{person}{MyungJoo Ham}, \bibinfo{person}{Jijoong Moon}, \bibinfo{person}{Geunsik Lim}, \bibinfo{person}{Jaeyun Jung}, \bibinfo{person}{Hyoungjoo Ahn}, \bibinfo{person}{Wook Song}, \bibinfo{person}{Sangjung Woo}, \bibinfo{person}{Parichay Kapoor}, \bibinfo{person}{Dongju Chae}, \bibinfo{person}{Gichan Jang}, \bibinfo{person}{Yongjoo Ahn}, {and} \bibinfo{person}{Jihoon Lee}.} \bibinfo{year}{2021}\natexlab{}.
\newblock \showarticletitle{{NNStreamer}: Efficient and Agile Development of On-Device AI Systems}. In \bibinfo{booktitle}{\emph{2021 IEEE/ACM 43rd International Conference on Software Engineering: Software Engineering in Practice (ICSE-SEIP)}}. IEEE, \bibinfo{pages}{198--207}.
\newblock


\bibitem[He et~al\mbox{.}(2016)]%
        {he2016deep}
\bibfield{author}{\bibinfo{person}{Kaiming He}, \bibinfo{person}{Xiangyu Zhang}, \bibinfo{person}{Shaoqing Ren}, {and} \bibinfo{person}{Jian Sun}.} \bibinfo{year}{2016}\natexlab{}.
\newblock \showarticletitle{Deep residual learning for image recognition}. In \bibinfo{booktitle}{\emph{Proceedings of the IEEE conference on computer vision and pattern recognition}}. \bibinfo{pages}{770--778}.
\newblock


\bibitem[Hinton et~al\mbox{.}(2012)]%
        {hinton2012deep}
\bibfield{author}{\bibinfo{person}{Geoffrey Hinton}, \bibinfo{person}{Li Deng}, \bibinfo{person}{Dong Yu}, \bibinfo{person}{George~E Dahl}, \bibinfo{person}{Abdel-rahman Mohamed}, \bibinfo{person}{Navdeep Jaitly}, \bibinfo{person}{Andrew Senior}, \bibinfo{person}{Vincent Vanhoucke}, \bibinfo{person}{Patrick Nguyen}, \bibinfo{person}{Tara~N Sainath}, {et~al\mbox{.}}} \bibinfo{year}{2012}\natexlab{}.
\newblock \showarticletitle{Deep neural networks for acoustic modeling in speech recognition: The shared views of four research groups}.
\newblock \bibinfo{journal}{\emph{IEEE Signal processing magazine}} \bibinfo{volume}{29}, \bibinfo{number}{6} (\bibinfo{year}{2012}), \bibinfo{pages}{82--97}.
\newblock


\bibitem[Huang et~al\mbox{.}(2019)]%
        {huang2019gpipe}
\bibfield{author}{\bibinfo{person}{Yanping Huang}, \bibinfo{person}{Youlong Cheng}, \bibinfo{person}{Ankur Bapna}, \bibinfo{person}{Orhan Firat}, \bibinfo{person}{Dehao Chen}, \bibinfo{person}{Mia Chen}, \bibinfo{person}{HyoukJoong Lee}, \bibinfo{person}{Jiquan Ngiam}, \bibinfo{person}{Quoc~V Le}, \bibinfo{person}{Yonghui Wu}, {et~al\mbox{.}}} \bibinfo{year}{2019}\natexlab{}.
\newblock \showarticletitle{Gpipe: Efficient training of giant neural networks using pipeline parallelism}.
\newblock \bibinfo{journal}{\emph{Advances in neural information processing systems}}  \bibinfo{volume}{32} (\bibinfo{year}{2019}).
\newblock


\bibitem[Jacob et~al\mbox{.}(2018)]%
        {jacob2018quantization}
\bibfield{author}{\bibinfo{person}{Benoit Jacob}, \bibinfo{person}{Skirmantas Kligys}, \bibinfo{person}{Bo Chen}, \bibinfo{person}{Menglong Zhu}, \bibinfo{person}{Matthew Tang}, \bibinfo{person}{Andrew Howard}, \bibinfo{person}{Hartwig Adam}, {and} \bibinfo{person}{Dmitry Kalenichenko}.} \bibinfo{year}{2018}\natexlab{}.
\newblock \showarticletitle{Quantization and training of neural networks for efficient integer-arithmetic-only inference}. In \bibinfo{booktitle}{\emph{Proceedings of the IEEE conference on computer vision and pattern recognition}}. \bibinfo{pages}{2704--2713}.
\newblock


\bibitem[Jia et~al\mbox{.}(2014)]%
        {jia2014caffe}
\bibfield{author}{\bibinfo{person}{Yangqing Jia}, \bibinfo{person}{Evan Shelhamer}, \bibinfo{person}{Jeff Donahue}, \bibinfo{person}{Sergey Karayev}, \bibinfo{person}{Jonathan Long}, \bibinfo{person}{Ross Girshick}, \bibinfo{person}{Sergio Guadarrama}, {and} \bibinfo{person}{Trevor Darrell}.} \bibinfo{year}{2014}\natexlab{}.
\newblock \showarticletitle{Caffe: Convolutional architecture for fast feature embedding}. In \bibinfo{booktitle}{\emph{Proceedings of the 22nd ACM international conference on Multimedia}}. \bibinfo{pages}{675--678}.
\newblock


\bibitem[{Justin Basilico}(2021)]%
        {netflix_personalization}
\bibfield{author}{\bibinfo{person}{{Justin Basilico}}.} \bibinfo{year}{2021}\natexlab{}.
\newblock \bibinfo{title}{Netflix Explains Recommendations and Personalization}.
\newblock \bibinfo{howpublished}{\url{http:https://scale.com/blog/Netflix-Recommendation-Personalization-TransformX-Scale-AI-Insights}}.
\newblock


\bibitem[Ketkar(2017)]%
        {ketkar2017stochastic}
\bibfield{author}{\bibinfo{person}{Nikhil Ketkar}.} \bibinfo{year}{2017}\natexlab{}.
\newblock \showarticletitle{Stochastic gradient descent}.
\newblock In \bibinfo{booktitle}{\emph{Deep learning with Python}}. \bibinfo{publisher}{Springer}, \bibinfo{pages}{113--132}.
\newblock


\bibitem[Krizhevsky et~al\mbox{.}(2017)]%
        {krizhevsky2017imagenet}
\bibfield{author}{\bibinfo{person}{Alex Krizhevsky}, \bibinfo{person}{Ilya Sutskever}, {and} \bibinfo{person}{Geoffrey~E Hinton}.} \bibinfo{year}{2017}\natexlab{}.
\newblock \showarticletitle{Imagenet classification with deep convolutional neural networks}.
\newblock \bibinfo{journal}{\emph{Commun. ACM}} \bibinfo{volume}{60}, \bibinfo{number}{6} (\bibinfo{year}{2017}), \bibinfo{pages}{84--90}.
\newblock


\bibitem[Li et~al\mbox{.}(2020)]%
        {li2020review}
\bibfield{author}{\bibinfo{person}{Li Li}, \bibinfo{person}{Yuxi Fan}, \bibinfo{person}{Mike Tse}, {and} \bibinfo{person}{Kuo-Yi Lin}.} \bibinfo{year}{2020}\natexlab{}.
\newblock \showarticletitle{A review of applications in federated learning}.
\newblock \bibinfo{journal}{\emph{Computers \& Industrial Engineering}} (\bibinfo{year}{2020}), \bibinfo{pages}{106854}.
\newblock


\bibitem[Lim et~al\mbox{.}(2021)]%
        {lim2021lightsys}
\bibfield{author}{\bibinfo{person}{Geunsik Lim}, \bibinfo{person}{MyungJoo Ham}, \bibinfo{person}{Jijoong Moon}, {and} \bibinfo{person}{Wook Song}.} \bibinfo{year}{2021}\natexlab{}.
\newblock \showarticletitle{{LightSys}: lightweight and efficient CI system for improving integration speed of software}. In \bibinfo{booktitle}{\emph{2021 IEEE/ACM 43rd International Conference on Software Engineering: Software Engineering in Practice (ICSE-SEIP)}}. IEEE, \bibinfo{pages}{1--10}.
\newblock


\bibitem[Lim et~al\mbox{.}(2023)]%
        {lim2023mobicom}
\bibfield{author}{\bibinfo{person}{Geunsik Lim}, \bibinfo{person}{Donghyun Kang}, \bibinfo{person}{MyungJoo Ham}, {and} \bibinfo{person}{Young~Ik Eom}.} \bibinfo{year}{2023}\natexlab{}.
\newblock \showarticletitle{SWAM: Revisiting Swap and OOMK for Improving Application Responsiveness on Mobile Devices}. In \bibinfo{booktitle}{\emph{MobiCom 2023 (Annual International Conference On Mobile Computing And Networking), To appear}}.
\newblock


\bibitem[Ling et~al\mbox{.}(2015)]%
        {ling2015deep}
\bibfield{author}{\bibinfo{person}{Zhen-Hua Ling}, \bibinfo{person}{Shi-Yin Kang}, \bibinfo{person}{Heiga Zen}, \bibinfo{person}{Andrew Senior}, \bibinfo{person}{Mike Schuster}, \bibinfo{person}{Xiao-Jun Qian}, \bibinfo{person}{Helen~M Meng}, {and} \bibinfo{person}{Li Deng}.} \bibinfo{year}{2015}\natexlab{}.
\newblock \showarticletitle{Deep learning for acoustic modeling in parametric speech generation: A systematic review of existing techniques and future trends}.
\newblock \bibinfo{journal}{\emph{IEEE Signal Processing Magazine}} \bibinfo{volume}{32}, \bibinfo{number}{3} (\bibinfo{year}{2015}), \bibinfo{pages}{35--52}.
\newblock


\bibitem[Lu et~al\mbox{.}(2015)]%
        {lu2015transfer}
\bibfield{author}{\bibinfo{person}{Jie Lu}, \bibinfo{person}{Vahid Behbood}, \bibinfo{person}{Peng Hao}, \bibinfo{person}{Hua Zuo}, \bibinfo{person}{Shan Xue}, {and} \bibinfo{person}{Guangquan Zhang}.} \bibinfo{year}{2015}\natexlab{}.
\newblock \showarticletitle{Transfer learning using computational intelligence: A survey}.
\newblock \bibinfo{journal}{\emph{Knowledge-Based Systems}}  \bibinfo{volume}{80} (\bibinfo{year}{2015}), \bibinfo{pages}{14--23}.
\newblock


\bibitem[{MIT Technical Review}({[n.\,d.]})]%
        {MITTechReview}
\bibfield{author}{\bibinfo{person}{{MIT Technical Review}}.} \bibinfo{year}{[n.\,d.]}\natexlab{}.
\newblock \bibinfo{title}{On-Device AI}.
\newblock \bibinfo{howpublished}{\url{https://www.technologyreview.com/hub/ubiquitous-on-device-ai/}}.
\newblock
\newblock
\shownote{(accessed 14 Dec 2021)}.


\bibitem[Mostafa and Wang(2019)]%
        {mostafa2019parameter}
\bibfield{author}{\bibinfo{person}{Hesham Mostafa} {and} \bibinfo{person}{Xin Wang}.} \bibinfo{year}{2019}\natexlab{}.
\newblock \showarticletitle{Parameter efficient training of deep convolutional neural networks by dynamic sparse reparameterization}. In \bibinfo{booktitle}{\emph{International Conference on Machine Learning}}. PMLR, \bibinfo{pages}{4646--4655}.
\newblock


\bibitem[Narang et~al\mbox{.}(2018)]%
        {narang2018mixed}
\bibfield{author}{\bibinfo{person}{Sharan Narang}, \bibinfo{person}{Gregory Diamos}, \bibinfo{person}{Erich Elsen}, \bibinfo{person}{Paulius Micikevicius}, \bibinfo{person}{Jonah Alben}, \bibinfo{person}{David Garcia}, \bibinfo{person}{Boris Ginsburg}, \bibinfo{person}{Michael Houston}, \bibinfo{person}{Oleksii Kuchaiev}, \bibinfo{person}{Ganesh Venkatesh}, {et~al\mbox{.}}} \bibinfo{year}{2018}\natexlab{}.
\newblock \showarticletitle{Mixed precision training}. In \bibinfo{booktitle}{\emph{Proc. 6th Int. Conf. on Learning Representations (ICLR)}}.
\newblock


\bibitem[{ndevilla}(2017)]%
        {iniparser}
\bibfield{author}{\bibinfo{person}{{ndevilla}}.} \bibinfo{year}{2017}\natexlab{}.
\newblock \bibinfo{title}{Iniparser4}.
\newblock \bibinfo{howpublished}{\url{https://github.com/ndevilla/iniparser}}.
\newblock


\bibitem[Pal and Pal(1993)]%
        {pal1993review}
\bibfield{author}{\bibinfo{person}{Nikhil~R Pal} {and} \bibinfo{person}{Sankar~K Pal}.} \bibinfo{year}{1993}\natexlab{}.
\newblock \showarticletitle{A review on image segmentation techniques}.
\newblock \bibinfo{journal}{\emph{Pattern recognition}} \bibinfo{volume}{26}, \bibinfo{number}{9} (\bibinfo{year}{1993}), \bibinfo{pages}{1277--1294}.
\newblock


\bibitem[Pascanu et~al\mbox{.}(2013)]%
        {pascanu2013difficulty}
\bibfield{author}{\bibinfo{person}{Razvan Pascanu}, \bibinfo{person}{Tomas Mikolov}, {and} \bibinfo{person}{Yoshua Bengio}.} \bibinfo{year}{2013}\natexlab{}.
\newblock \showarticletitle{On the difficulty of training recurrent neural networks}. In \bibinfo{booktitle}{\emph{International conference on machine learning}}. PMLR, \bibinfo{pages}{1310--1318}.
\newblock


\bibitem[Paszke et~al\mbox{.}(2017)]%
        {paszke2017automatic}
\bibfield{author}{\bibinfo{person}{Adam Paszke}, \bibinfo{person}{Sam Gross}, \bibinfo{person}{Soumith Chintala}, \bibinfo{person}{Gregory Chanan}, \bibinfo{person}{Edward Yang}, \bibinfo{person}{Zachary DeVito}, \bibinfo{person}{Zeming Lin}, \bibinfo{person}{Alban Desmaison}, \bibinfo{person}{Luca Antiga}, {and} \bibinfo{person}{Adam Lerer}.} \bibinfo{year}{2017}\natexlab{}.
\newblock \showarticletitle{Automatic differentiation in pytorch}. In \bibinfo{booktitle}{\emph{NIPS 2017 Workshop on Autodiff}} (Long Beach, California, USA).
\newblock
\urldef\tempurl%
\url{https://openreview.net/forum?id=BJJsrmfCZ}
\showURL{%
\tempurl}


\bibitem[Qian et~al\mbox{.}(2014)]%
        {qian2014training}
\bibfield{author}{\bibinfo{person}{Yao Qian}, \bibinfo{person}{Yuchen Fan}, \bibinfo{person}{Wenping Hu}, {and} \bibinfo{person}{Frank~K Soong}.} \bibinfo{year}{2014}\natexlab{}.
\newblock \showarticletitle{On the training aspects of deep neural network (DNN) for parametric TTS synthesis}. In \bibinfo{booktitle}{\emph{2014 IEEE International Conference on Acoustics, Speech and Signal Processing (ICASSP)}}. IEEE, \bibinfo{pages}{3829--3833}.
\newblock


\bibitem[Rajbhandari et~al\mbox{.}(2021)]%
        {rajbhandari2021zero}
\bibfield{author}{\bibinfo{person}{Samyam Rajbhandari}, \bibinfo{person}{Olatunji Ruwase}, \bibinfo{person}{Jeff Rasley}, \bibinfo{person}{Shaden Smith}, {and} \bibinfo{person}{Yuxiong He}.} \bibinfo{year}{2021}\natexlab{}.
\newblock \showarticletitle{Zero-infinity: Breaking the gpu memory wall for extreme scale deep learning}. In \bibinfo{booktitle}{\emph{Proceedings of the International Conference for High Performance Computing, Networking, Storage and Analysis}}. \bibinfo{pages}{1--14}.
\newblock


\bibitem[Ren et~al\mbox{.}(2021)]%
        {ren2021zero}
\bibfield{author}{\bibinfo{person}{Jie Ren}, \bibinfo{person}{Samyam Rajbhandari}, \bibinfo{person}{Reza~Yazdani Aminabadi}, \bibinfo{person}{Olatunji Ruwase}, \bibinfo{person}{Shuangyan Yang}, \bibinfo{person}{Minjia Zhang}, \bibinfo{person}{Dong Li}, {and} \bibinfo{person}{Yuxiong He}.} \bibinfo{year}{2021}\natexlab{}.
\newblock \showarticletitle{$\{$ZeRO-Offload$\}$: Democratizing $\{$Billion-Scale$\}$ Model Training}. In \bibinfo{booktitle}{\emph{2021 USENIX Annual Technical Conference (USENIX ATC 21)}}. \bibinfo{pages}{551--564}.
\newblock


\bibitem[Shah et~al\mbox{.}(2020)]%
        {shah2020memory}
\bibfield{author}{\bibinfo{person}{Aashaka Shah}, \bibinfo{person}{Chao-Yuan Wu}, \bibinfo{person}{Jayashree Mohan}, \bibinfo{person}{Vijay Chidambaram}, {and} \bibinfo{person}{Philipp Kr{\"a}henb{\"u}hl}.} \bibinfo{year}{2020}\natexlab{}.
\newblock \showarticletitle{Memory optimization for deep networks}.
\newblock \bibinfo{journal}{\emph{arXiv preprint arXiv:2010.14501}} (\bibinfo{year}{2020}).
\newblock


\bibitem[Shen et~al\mbox{.}(2018)]%
        {shen2018natural}
\bibfield{author}{\bibinfo{person}{Jonathan Shen}, \bibinfo{person}{Ruoming Pang}, \bibinfo{person}{Ron~J Weiss}, \bibinfo{person}{Mike Schuster}, \bibinfo{person}{Navdeep Jaitly}, \bibinfo{person}{Zongheng Yang}, \bibinfo{person}{Zhifeng Chen}, \bibinfo{person}{Yu Zhang}, \bibinfo{person}{Yuxuan Wang}, \bibinfo{person}{Rj Skerrv-Ryan}, {et~al\mbox{.}}} \bibinfo{year}{2018}\natexlab{}.
\newblock \showarticletitle{Natural tts synthesis by conditioning wavenet on mel spectrogram predictions}. In \bibinfo{booktitle}{\emph{2018 IEEE International Conference on Acoustics, Speech and Signal Processing (ICASSP)}}. IEEE, \bibinfo{pages}{4779--4783}.
\newblock


\bibitem[Shin et~al\mbox{.}(2017)]%
        {shin2017continual}
\bibfield{author}{\bibinfo{person}{Hanul Shin}, \bibinfo{person}{Jung~Kwon Lee}, \bibinfo{person}{Jaehong Kim}, {and} \bibinfo{person}{Jiwon Kim}.} \bibinfo{year}{2017}\natexlab{}.
\newblock \showarticletitle{Continual learning with deep generative replay}.
\newblock \bibinfo{journal}{\emph{arXiv preprint arXiv:1705.08690}} (\bibinfo{year}{2017}).
\newblock


\bibitem[Simonyan and Zisserman(2014)]%
        {simonyan2014very}
\bibfield{author}{\bibinfo{person}{Karen Simonyan} {and} \bibinfo{person}{Andrew Zisserman}.} \bibinfo{year}{2014}\natexlab{}.
\newblock \showarticletitle{Very deep convolutional networks for large-scale image recognition}.
\newblock \bibinfo{journal}{\emph{arXiv preprint arXiv:1409.1556}} (\bibinfo{year}{2014}).
\newblock


\bibitem[Tan et~al\mbox{.}(2018)]%
        {tan2018survey}
\bibfield{author}{\bibinfo{person}{Chuanqi Tan}, \bibinfo{person}{Fuchun Sun}, \bibinfo{person}{Tao Kong}, \bibinfo{person}{Wenchang Zhang}, \bibinfo{person}{Chao Yang}, {and} \bibinfo{person}{Chunfang Liu}.} \bibinfo{year}{2018}\natexlab{}.
\newblock \showarticletitle{A survey on deep transfer learning}. In \bibinfo{booktitle}{\emph{International conference on artificial neural networks}}. Springer, \bibinfo{pages}{270--279}.
\newblock


\bibitem[Teerapittayanon et~al\mbox{.}(2016)]%
        {7900006}
\bibfield{author}{\bibinfo{person}{Surat Teerapittayanon}, \bibinfo{person}{Bradley McDanel}, {and} \bibinfo{person}{H.T. Kung}.} \bibinfo{year}{2016}\natexlab{}.
\newblock \showarticletitle{BranchyNet: Fast inference via early exiting from deep neural networks}. In \bibinfo{booktitle}{\emph{2016 23rd International Conference on Pattern Recognition (ICPR)}}. \bibinfo{pages}{2464--2469}.
\newblock
\urldef\tempurl%
\url{https://doi.org/10.1109/ICPR.2016.7900006}
\showDOI{\tempurl}


\bibitem[Thrun and Mitchell(1995)]%
        {thrun1995lifelong}
\bibfield{author}{\bibinfo{person}{Sebastian Thrun} {and} \bibinfo{person}{Tom~M Mitchell}.} \bibinfo{year}{1995}\natexlab{}.
\newblock \showarticletitle{Lifelong robot learning}.
\newblock \bibinfo{journal}{\emph{Robotics and autonomous systems}} \bibinfo{volume}{15}, \bibinfo{number}{1-2} (\bibinfo{year}{1995}), \bibinfo{pages}{25--46}.
\newblock


\bibitem[Torrey and Shavlik(2010)]%
        {torrey2010transfer}
\bibfield{author}{\bibinfo{person}{Lisa Torrey} {and} \bibinfo{person}{Jude Shavlik}.} \bibinfo{year}{2010}\natexlab{}.
\newblock \showarticletitle{Transfer learning}.
\newblock In \bibinfo{booktitle}{\emph{Handbook of research on machine learning applications and trends: algorithms, methods, and techniques}}. \bibinfo{publisher}{IGI global}, \bibinfo{pages}{242--264}.
\newblock


\bibitem[Vaswani et~al\mbox{.}(2017)]%
        {vaswani2017attention}
\bibfield{author}{\bibinfo{person}{Ashish Vaswani}, \bibinfo{person}{Noam Shazeer}, \bibinfo{person}{Niki Parmar}, \bibinfo{person}{Jakob Uszkoreit}, \bibinfo{person}{Llion Jones}, \bibinfo{person}{Aidan~N Gomez}, \bibinfo{person}{{\L}ukasz Kaiser}, {and} \bibinfo{person}{Illia Polosukhin}.} \bibinfo{year}{2017}\natexlab{}.
\newblock \showarticletitle{Attention is all you need}.
\newblock \bibinfo{journal}{\emph{Advances in neural information processing systems}}  \bibinfo{volume}{30} (\bibinfo{year}{2017}).
\newblock


\bibitem[Voigt and Von~dem Bussche(2017)]%
        {voigt2017eu}
\bibfield{author}{\bibinfo{person}{Paul Voigt} {and} \bibinfo{person}{Axel Von~dem Bussche}.} \bibinfo{year}{2017}\natexlab{}.
\newblock \showarticletitle{The eu general data protection regulation (gdpr)}.
\newblock \bibinfo{journal}{\emph{A Practical Guide, 1st Ed., Cham: Springer International Publishing}}  \bibinfo{volume}{10} (\bibinfo{year}{2017}), \bibinfo{pages}{3152676}.
\newblock


\bibitem[Williams and Zipser(1989)]%
        {williams1989learning}
\bibfield{author}{\bibinfo{person}{Ronald~J Williams} {and} \bibinfo{person}{David Zipser}.} \bibinfo{year}{1989}\natexlab{}.
\newblock \showarticletitle{A learning algorithm for continually running fully recurrent neural networks}.
\newblock \bibinfo{journal}{\emph{Neural computation}} \bibinfo{volume}{1}, \bibinfo{number}{2} (\bibinfo{year}{1989}), \bibinfo{pages}{270--280}.
\newblock


\bibitem[Wu et~al\mbox{.}(2015)]%
        {wu2015deep}
\bibfield{author}{\bibinfo{person}{Ren Wu}, \bibinfo{person}{Shengen Yan}, \bibinfo{person}{Yi Shan}, \bibinfo{person}{Qingqing Dang}, {and} \bibinfo{person}{Gang Sun}.} \bibinfo{year}{2015}\natexlab{}.
\newblock \showarticletitle{Deep image: Scaling up image recognition}.
\newblock \bibinfo{journal}{\emph{arXiv preprint arXiv:1501.02876}} \bibinfo{volume}{7}, \bibinfo{number}{8} (\bibinfo{year}{2015}).
\newblock


\bibitem[Yang et~al\mbox{.}(2019)]%
        {yang2019deep}
\bibfield{author}{\bibinfo{person}{Wenming Yang}, \bibinfo{person}{Xuechen Zhang}, \bibinfo{person}{Yapeng Tian}, \bibinfo{person}{Wei Wang}, \bibinfo{person}{Jing-Hao Xue}, {and} \bibinfo{person}{Qingmin Liao}.} \bibinfo{year}{2019}\natexlab{}.
\newblock \showarticletitle{Deep learning for single image super-resolution: A brief review}.
\newblock \bibinfo{journal}{\emph{IEEE Transactions on Multimedia}} \bibinfo{volume}{21}, \bibinfo{number}{12} (\bibinfo{year}{2019}), \bibinfo{pages}{3106--3121}.
\newblock


\bibitem[Zhang et~al\mbox{.}(2020)]%
        {zhang2020improved}
\bibfield{author}{\bibinfo{person}{Bohang Zhang}, \bibinfo{person}{Jikai Jin}, \bibinfo{person}{Cong Fang}, {and} \bibinfo{person}{Liwei Wang}.} \bibinfo{year}{2020}\natexlab{}.
\newblock \showarticletitle{Improved analysis of clipping algorithms for non-convex optimization}.
\newblock \bibinfo{journal}{\emph{Advances in Neural Information Processing Systems}}  \bibinfo{volume}{33} (\bibinfo{year}{2020}), \bibinfo{pages}{15511--15521}.
\newblock


\bibitem[{Zhang Xianyi}(2013)]%
        {blas}
\bibfield{author}{\bibinfo{person}{{Zhang Xianyi}}.} \bibinfo{year}{2013}\natexlab{}.
\newblock \bibinfo{title}{OpenBLAS}.
\newblock \bibinfo{howpublished}{\url{http:https://github.com/xianyi/OpenBLAS}}.
\newblock


\end{thebibliography}

\end{document}